\documentclass{article}

\PassOptionsToPackage{numbers, compress}{natbib}

\usepackage[preprint]{neurips_2026}

\usepackage[utf8]{inputenc}
\usepackage[T1]{fontenc}
\usepackage{url}
\usepackage{booktabs}
\usepackage{amsfonts}
\usepackage{nicefrac}
\usepackage{microtype}
\usepackage{xcolor}

\definecolor{neuripsblue}{rgb}{0.21,0.49,0.74}
\definecolor{darkgreen}{rgb}{0.0, 0.5, 0.0}
\usepackage[breaklinks,colorlinks,citecolor=neuripsblue]{hyperref}

\usepackage{graphicx}
\usepackage{subcaption}
\usepackage{wrapfig}

\usepackage{amsmath}
\usepackage{amssymb}
\usepackage{mathtools}
\usepackage{amsthm}
\usepackage{multirow}
\usepackage{makecell}

\usepackage[capitalize,noabbrev]{cleveref}

\theoremstyle{plain}

\theoremstyle{definition}

\theoremstyle{remark}

\usepackage[textsize=tiny]{todonotes}

\begin{document}

\title{Restoration-Aligned Generative Flow Models for\\Blind Motion Deblurring}

\author{
  Insoo Kim$^{1}$\thanks{This work was done at Samsung Electronics and KAIST AI.} \\
  $^{1}$NAVER Cloud
  \And
  Jinwoo Shin$^{2}$\thanks{Corresponding author.} \\
  $^{2}$KAIST AI
}

\maketitle

\begin{abstract}
Generative flow models offer powerful priors learned from large-scale natural images, but directly adapting them to restoration tasks such as motion deblurring causes severe fidelity degradation, as their training objective is inherently misaligned with restoration. We present DeblurFlow, a framework that resolves this misalignment by reformulating the flow trajectory itself: we replace the noise endpoint with the blur observation, which makes the underlying vector field coincide with the residual error between blur and clean images. Under this formulation, the standard flow matching loss naturally takes the form of a residual loss, allowing pretrained flow models to be optimized under restoration-aligned objectives via LoRA adaptation. This formulation further enables a dual-expert sampling strategy: a fidelity expert provides a high-fidelity initialization, e.g., PSNR 33.69 dB, and DeblurFlow enhances perceptual quality with only a marginal fidelity reduction to 33.05 dB, whereas directly applying a generative model on top of a fidelity expert decreases PSNR to 27.60 dB. To make this practical, we further introduce r-space, a latent space tailored for residual decoding rather than image reconstruction, which reduces encoder–decoder cost by up to 9× over standard VAE latents. Extensive experiments on GoPro, HIDE, RealBlur, and RWBI demonstrate that DeblurFlow achieves strong restoration fidelity and perceptual realism, while remaining computationally practical.
\end{abstract}

\section{Introduction}
\label{sec:intro}
Motion blur is a common degradation in photography, arising from camera shakes and object movements in long exposure times. Blind motion deblurring aims to recover a clean latent image from a single blur observation.
In recent years, diverse deep restoration architectures have shown significant advances in image deblurring~\cite{gopro,deblurganv2,mimounet,mprnet,maxim,msdi,nafnet,restormer,uformer,stripformer,fftformer,ufp,segdeblur,adarev, geosyn, deblurdiff, fidediff}, allowing for high-fidelity reconstructions that closely align with ground-truth images.
They are typically optimized with pixel-wise losses, which often struggle to produce perceptually realistic details~\cite{tradeoff}.
In contrast, diffusion~\cite{ddpm,ddim,score,stablediff,pixart} and flow matching models~\cite{flowmodel,instaflow,recflow,sana} introduce a generative paradigm that enables high-quality image synthesis conditioned on text, semantic cues, or degraded images.
These two paradigms represent complementary strengths: restoration-based methods~\cite{restormer,nafnet,uformer,maxim} excel at restoration fidelity while generative methods\cite{stablediff,pixart,flowmodel,sana}, equipped with powerful priors learned from large-scale clean images~\cite{laion,coyo}, excel at perceptual quality.

In practice, combining these two strengths has proven fundamentally challenging, as their objectives are inherently incompatible. Generative flow models are trained to transport noise to clean image distributions. Applying this generation process to deblurring tasks struggles with poor fidelity (e.g., low PSNR) or fails to preserve even high-fidelity initialization (e.g., a significant PSNR drop) due to their generation-oriented objectives, i.e., flow matching loss, rather than restoration-oriented objectives. The key question is how to reformulate the trajectory so that the generative flow model becomes compatible with deblurring tasks.

\begin{figure}[t]
  \centering
  \begin{minipage}[t]{0.49\linewidth}
    \centering
    \includegraphics[width=\linewidth]{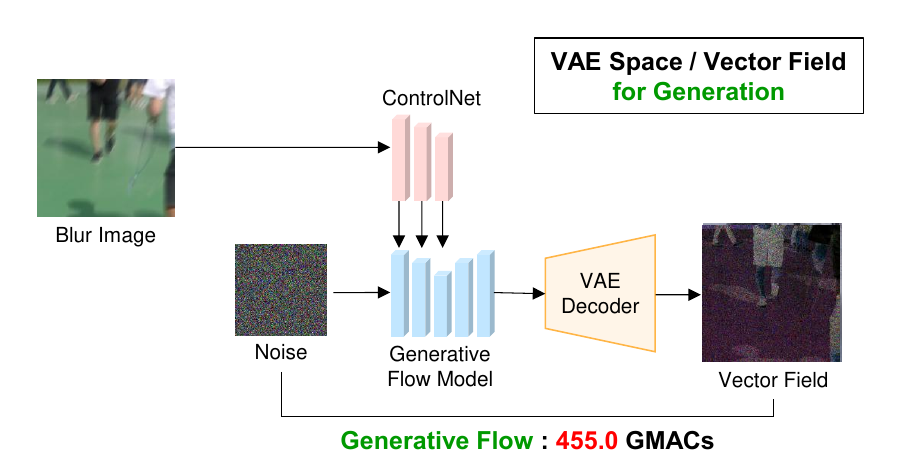}
  \end{minipage}\hfill
  \begin{minipage}[t]{0.49\linewidth}
    \centering
    \includegraphics[width=\linewidth]{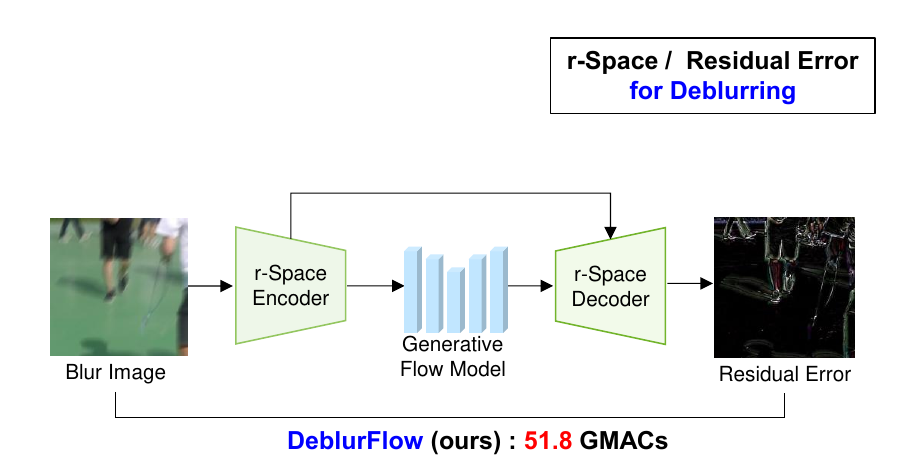}
  \end{minipage}
  \vspace{-0.1cm}
  \caption{Main concept of our DeblurFlow. Unlike conditional generative flow models that generate vector fields for deblurring, our method produces task-aligned residuals using a r-space encoder–decoder tailored for restoration, notably reducing GMACs.}
  \label{fig:concept_proposed}
\vspace{-0.3cm}
\end{figure}

In this paper, we propose DeblurFlow, a generative deblurring framework that reformulates the generative flow trajectory from a restoration perspective.
Rather than treating the generative vector field as a transport from noise to clean, we reinterpret it as a residual error: by replacing the noise endpoint with the blur observation, the underlying vector field naturally becomes the difference between blur and clean images, i.e., residual error, as shown in Fig.~\ref{fig:concept_proposed}, and the flow matching objective~\cite{flowmodel} coincides with the residual loss~\cite{segdeblur}. This enables flow-based generative frameworks to be trained under deblurring-oriented objectives via Low-Rank Adaptation (LoRA) adaptation~\cite{lora}. As a result, such restoration alignment empowers DeblurFlow to naturally act as a fidelity-preserving generative deblurring model.

\begin{figure*}[!t]
  \centering
  \centerline{\includegraphics[width=14.4cm]{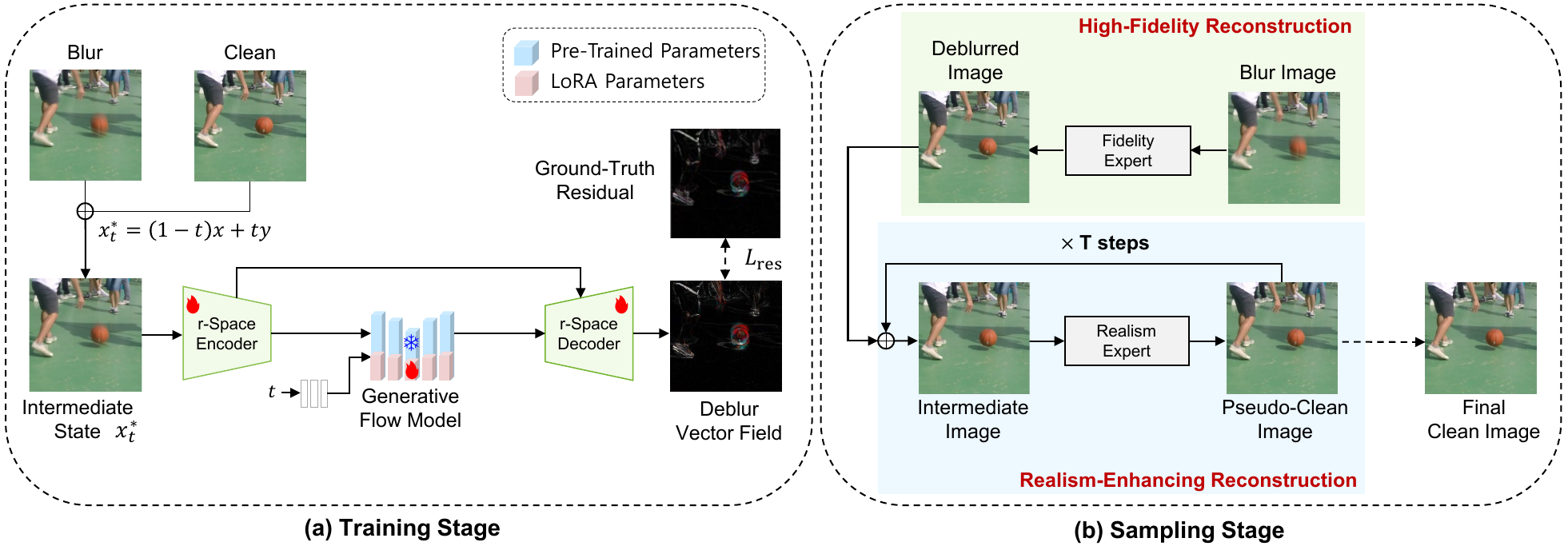}}
  \vspace{-0.1cm}
  \caption{Overview of the proposed DeblurFlow framework. Our DeblurFlow produces a deblurring-oriented vector field, i.e., residual error, which is preferred for deblurring tasks. The r-space encoder-decoder, connected by the residual shortcuts, enables fidelity-preserving residual learning while reducing computational cost from 422.7 to 19.5 GMACs compared to the VAE encoder–decoder.} 
  \label{fig:proposed_method}
\vspace{-0.3cm}
\end{figure*}

This property motivates a dual-expert sampling strategy: rather than forcing a single model to balance restoration fidelity and perceptual quality simultaneously, we decompose the problem into two complementary roles, handled by a fidelity expert and a fidelity-preserving realism expert. A fidelity expert, e.g., a pretrained deblurring network~\cite{nafnet,fftformer}, first provides a high-fidelity initialization, and our DeblurFlow serves as a fidelity-preserving realism expert that enhances perceptual quality while mitigating the fidelity collapse, as shown in Fig.~\ref{fig:proposed_method} (b).
Concretely, naively applying a generative deblurring model on top of a fidelity expert without our formulation causes fidelity collapse (PSNR 33.69 → 27.60 dB), whereas our restoration-aligned formulation preserves competitive fidelity (33.05 dB) while substantially improving perceptual realism (MUSIQ 45.28 → 50.81).

To make this formulation practical, we further introduce r-space, a restoration-oriented latent space tailored for residual decoding rather than clean image reconstruction, as shown in Fig.~\ref{fig:proposed_method} (a). Standard VAE latent spaces are designed for reconstructing clean images, which limits their effectiveness for residual prediction. By tailoring our r-space for residual decoding and integrating skip connections that propagate structural cues directly from the blur input, we achieve more efficient fidelity-aware learning, reducing the computational cost from 455.0 to 51.8 GMACs, as shown in Fig.~\ref{fig:concept_proposed}.

We demonstrate the superiority of our method on GoPro~\cite{gopro}, HIDE~\cite{hide}, RealBlur~\cite{realblur} and RWBI~\cite{dbgan}. Our DeblurFlow consistently achieves strong restoration fidelity and perceptual realism across all benchmarks, while processing 2K resolution images in 0.41 seconds on a single GPU, approximately 3.4× faster than DiffIR~\cite{diffir} and 6.5× faster than FFTFormer~\cite{fftformer}. Our contributions are threefold:
\vspace{-0.2cm}
\begin{itemize}
    \item We propose a new deblurring framework that resolves the objective mismatch between generation and restoration by the restoration-aligned formulation. This mitigates fidelity collapse and makes generative flow models compatible with high-fidelity deblurring models.
    \item We introduce a restoration-oriented latent space, i.e., r-space, that enables fidelity-preserving encoding and decoding with computational efficiency.
    \item We introduce a dual-expert sampling strategy that ensures high restoration fidelity and perceptual realism.
\end{itemize}


\section{Related Works}
\label{sec:related_works}
\setlength{\parindent}{0in}\textbf{Restoration-based deblurring methods.}
Over the years, the restoration-based deblurring methods~\cite{mimounet,mprnet,maxim,nafnet,fftformer,ufp,segdeblur,refdeblur} have evolved along various aspects. Early studies utilize hierarchical recovery, where coarse-to-fine pipelines progressively restore fine details across multiple scales~\cite{gopro,srndeblurnet}. Another research line to adopt multi-input and multi-output U-Net variants~\cite{mimounet,deeprft,mprnet} is investigated to improve efficiency. More recently, transformer-based architectures~\cite{restormer,uformer,stripformer,fftformer,grl} have been explored, aiming to capture global dependencies and long-range context.
Some recent efforts have incorporated auxiliary priors~\cite{msdi,ufp,segdeblur,refdeblur} to enhance perceptual quality.
These restoration-driven methods deliver strong restoration fidelity, but their predictions tend to be over-smoothed and often fail to recover realistic texture details. In contrast, our method is initialized with the output of these restoration methods, which secures fidelity. Then, it benefits from generative priors of generative models, captured from large-scale clean images, leading to better perceptual quality.

\setlength{\parindent}{0in}\textbf{Diffusion-based deblurring methods.}
Diffusion-based methods treat deblurring as a conditional generative modeling problem.
Early studies~\cite{palette,indi,i2sb,ddb} have demonstrated that they can produce sharper and more realistic deblurring results than traditional restoration-based methods by finetuning the pre-trained diffusion model with synthetic data. 
Some literature~\cite{diffir,hidiff} incorporates diffusion priors to guide restoration networks, but they struggle to fully exploit the powerful generative priors of diffusion models for perceptual realism. An alternative method~\cite {dvsr,irsde} that explores diffusion models for deblurring tasks demonstrates improved perceptual quality, but often sacrifices restoration fidelity due to the misalignment between their generation and deblurring objectives.
In contrast, our method repurposes a pre-trained flow matching model by aligning it with the deblurring objective through a task-specific residual loss, leading to fidelity preservation and realism enhancement within generative deblurring models.

\section{DeblurFlow: Restoration-Aligned Generative Deblurring Models}
\subsection{Rethinking Generative Models for Deblurring}\label{sec:preliminary}
\setlength{\parindent}{0.4cm}
The fundamental goal of motion deblurring is to recover a clean image $x$ from its blur counterpart $y$. A common and effective method in deblurring is to predict the residual error, defined as the difference between blur and clean images: $r = y - x$. Therefore, the deblurring objective using a restoration network $f_\psi$ parameterized by $\psi$ is to minimize the discrepancy between the predicted residual error $r_\text{pred} = {f_\psi}(y)$ and the ground-truth residual error $r$, leading to the residual loss:

\begin{equation} \label{eq:res_loss}
\mathcal{L}_\text{res}(\psi) = ||r_\text{pred} - r||^2 = ||{f_\psi}(y) - (y - x)||^2.
\end{equation}
This formulation enforces the model to remove the blur component and to preserve the underlying content of the input, leading to fidelity-aware learning.
In contrast, the flow matching model aims to learn a vector field that transports a sample from a simple prior distribution (e.g., Gaussian noise) to the target data distribution of clean images, as shown in Fig.~\ref{fig:y_to_x}. Unlike diffusion models that rely on stochastic sampling, the flow matching model constructs a deterministic trajectory, enabling efficient generation with fewer sampling steps.  
This can be realized by minimizing the following loss:

\begin{equation} \label{eq:flow_loss}
\mathcal{L}_{\text{flow}}(\theta) = \| v_\theta(x_t, t) - (\epsilon - x) \|^2,
\end{equation}
where $v_\theta(x_t, t)$ denotes the vector field predicted by the flow matching model at time $t$, $\epsilon \sim \mathcal{N}(0, I)$ is a noise sample, and $x_t$ is the path sample at time $t$ that forms a linear interpolation between the ground-truth clean image $x$ and a noise sample $\epsilon$: 

\begin{equation} \label{eq:flow_xt}
x_t = (1-t)x + t\epsilon.
\end{equation}

While the restoration-based deblurring methods exhibit strong deblurring capability and high-fidelity reconstruction, the generative-based deblurring models, based on flow matching or diffusion models, are particularly effective in producing photorealistic details. As existing methods often excel in only one dimension (e.g., distortion or perceptual quality), we assign complementary roles to the two paradigms, i.e., restoration-based and generative-based methods, to maximize both restoration fidelity or perceptual realism. This will be further discussed in the following sections.

\subsection{Repurposing Generative Flows for Deblurring}\label{sec:deblurflow}
\begin{wrapfigure}{r}{0.50\linewidth}
\vspace{-0.8em}
  \centering
  \includegraphics[width=\linewidth]{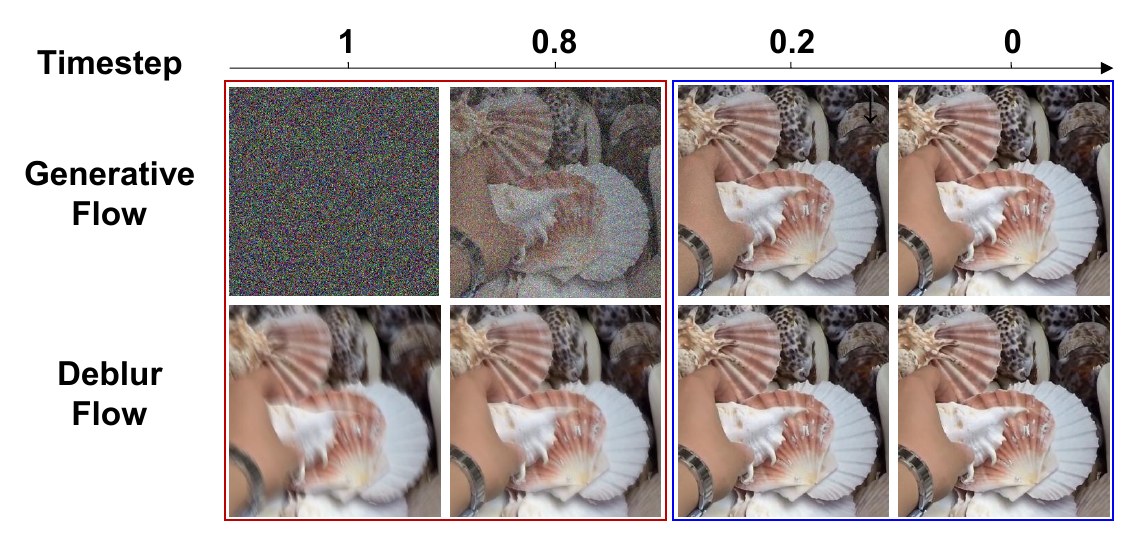}
  \vspace{-0.6em}
  \caption{Generative Flow vs. DeblurFlow.}
  \label{fig:y_to_x}
\vspace{-0.8em}
\end{wrapfigure}
When a generative model trained for deblurring tasks conducts even a single sampling step, it often causes a substantial drop in fidelity, i.e., low PSNR, because its training objective remains generation-oriented rather than restoration-oriented.
The key question is how to reformulate the trajectory so that the pretrained generative flow becomes compatible with deblurring. We replace the noise endpoint $\epsilon$ in~\eqref{eq:flow_xt} with the blur observation $y$, i.e., $y \rightarrow x$, as shown in Fig.~\ref{fig:y_to_x}, whose vector field naturally becomes a residual error $y - x$. This allows generative flow models to inherit the fidelity-preserving benefits of residual learning, thereby mitigating the fidelity degradation commonly observed in generative deblurring models~\cite{irsde,diffbir}. Meanwhile, it contributes to enhancing perceptual quality by leveraging its inherent generative priors.
To this end, we first describe the deblurring path sample $x_t^\star$ as

\begin{equation} \label{eq:deblurflow_xt}
x_t^\star = (1-t)\,x + t\,y,
\end{equation}
where $y$ is the blur input and $x$ is the clean ground truth. 
The deblurring-oriented vector field for this path is its derivative with respect to $t$:
\[
v^\star = \frac{d x_t}{dt} = y - x.
\]
This formulation naturally leads to a new task-aligned residual loss, based on the standard flow matching loss in~\eqref{eq:flow_loss}:

\begin{equation} \label{eq:deblurflow_loss1}
\mathcal{L}_{\text{res}}(\theta)
= \left\| v_\theta(x_t^\star,t) - (y - x) \right\|^2.
\end{equation}
{To leverage generative prior of the flow matching model, we introduce  Low-Rank Adaptation (LoRA)~\cite{lora}, parameterized by $\phi$ and train only this parameter to compensate for the distribution shift}, as discussed in Section~\ref{sec:preliminary}: 
\begin{equation} \label{eq:deblurflow_loss2}
\mathcal{L}_{\text{res}}(\phi)
  = \left\| v_{\theta,\phi}(x_t^\star, t) - (y - x) \right\|^2
\end{equation}
where $v_{\theta,\phi}(x_t^\star,t)$ denotes the domain-adapted residual prediction. 
The flow matching model is initially trained with noise-to-clean vector fields, causing domain gaps when applied to blur-to-clean vector fields. The LoRA parameters correct these mismatches, producing task-aligned residual errors.
We emphasize that this deblurring-oriented path contributes to making the generative flow model capable of mitigating fidelity drop, so that it can be integrated with a restoration-based model while minimizing fidelity degradation, thereby enabling a deblurring pipeline that achieves both restoration fidelity and perceptual realism.

\subsection{Latent DeblurFlow with r-Space}\label{sec:latent_rspace}
In the previous section, we discuss our generative deblurring framework, termed DeblurFlow, in the pixel space. In practice, we operate our DeblurFlow in the latent space for computational efficiency. {The standard VAE latent space~\cite{vae}, denoted as v-space, is not specifically designed for restoration; in particular, it is not tailored for image residual decoding, which limits its effectiveness for restoration tasks.}
To this end, as illustrated in Fig.~\ref{fig:proposed_method} (a), 
we propose a restoration-oriented latent space, termed r-space, by introducing a new encoder–decoder pair $\{\mathcal{E}_\varphi, \mathcal{D}_\varphi\}$ parameterized by $\varphi$.
Given a path sample $x_t^\star$, the r-space latent representation is $z_r = \mathcal{E}_\varphi(x_t^\star)$. The LoRA parameter $\phi$ defined by~\eqref{eq:deblurflow_loss2} plays an additional role in the latent space to align the v-space predictions with r-space ones, expressed as $v_{\theta,\phi}(z_r,t)$. Then, the r-space latent prediction is decoded by the r-space decoder $\mathcal{D}_\varphi$, to produce the image residual error, $\hat{r} = \mathcal{D}_\varphi(v_{\theta,\phi}(z_r,t))$. Finally, we suggest optimizing the following latent version of~\eqref{eq:deblurflow_loss2}: 
\begin{equation} \label{eq:deblurflow_loss3}
\mathcal{L}_{\text{res}}(\phi, \varphi)
= 
\left\| \hat{r} - (y - x) \right\|^2
\end{equation}

Here, we remark that the r-space provides the following benefits. It can leverage residual shortcuts, commonly used in restoration networks, to explicitly provide fidelity-preserving cues to the decoder, as shown in Fig.~\ref{fig:proposed_method} (a). This architectural design enables fidelity-aware deblurring beyond the capability of the standard VAE, and reduces computational costs, cutting the encoder–decoder computational cost from $422.7$ GMACs to $19.5$ GMACs\footnote{We compare our DeblurFlow with previous methods, analyzing both computational cost and performance, in Section~\ref{sec:ablation}.}, enabling it to be practical for real-world applications.

\subsection{Dual-Expert Sampling Strategy}\label{sec:proposed_sampling}
Existing deblurring methods often achieve strong performance in either fidelity or perceptual quality, but not both simultaneously. For example, generative deblurring methods offer strong perceptual quality, but they often suffer from severe fidelity collapse. Rather than forcing a single monolithic model to achieve both aspects simultaneously, we decompose the problem into two complementary roles, handled by a fidelity expert and a fidelity-preserving realism expert. This strategy is to push its boundary further by achieving a stronger balance across key metrics, as shown in Tables~\ref{tbl:gopro_merged} and~\ref{tbl:hide_realblur_merged}.
Our DeblurFlow is architecturally designed with a task-aligned residual loss to ensure fidelity-preserving reconstruction while contributing perceptual quality as discussed in Section~\ref{sec:deblurflow}.
A pre-trained deblurring model $f_\psi$ serves as the fidelity expert, providing a high-fidelity initial estimate $x_t^\star=f_\psi(y)$ from the blur observation $y$. Starting from this estimate, our DeblurFlow acts as the fidelity-preserving realism expert, refining the initial estimate through the predicted residual error $\mathcal{D}_\varphi(v_{\theta,\phi}(z_r,t))$ under the flow matching sampling process, improving perceptual quality while preserving the established fidelity, i.e.,
\begin{equation}\label{eq:sampling_residual}
    x_{t-\Delta t}^{\star} = x_t^\star - \mathcal{D}_\varphi(v_{\theta,\phi}(z_r,t))\Delta t,
\end{equation}
where $x_{t-\Delta t}^{\star}$ represents the updated sample after one refinement step and $\Delta t$ denotes the sampling step size along the time trajectory.
By repeating~\eqref{eq:sampling_residual} until $t=0$, the realism expert progressively improves its perceptual realism\footnote{We discuss the effect of sampling steps in Section~\ref{sec:app_n_steps}.}, ultimately achieving photorealistic deblurring.
{Under this sampling strategy, a potential distribution gap may arise if the model is trained only on the blur-to-clean mapping ($y \to x$), as it is trained on blur inputs rather than high-fidelity initializations. To resolve this issue, we adopt a co-training strategy that jointly learns from the full degradation mapping ($y \to x$) and the expert-dependent mapping ($f_\psi(y) \to x$) with sampling probabilities of 0.7 and 0.3, respectively. This dual training ensures domain alignment while maintaining the model's capability to prevent overfitting to specific fidelity experts and enable robustness to challenging motion blur beyond the coverage of fidelity experts, as shown in Fig.~\ref{fig:app_result_rwbi}. A detailed analysis is provided in Section~\ref{sec:app_discussion_using_fidelity_expert}.}

\subsection{Discussion}\label{sec:discussion}
\setlength{\parindent}{0in}\textbf{Role of our formulation.} 
Our contribution lies not in the components (i.e., fidelity and realism experts) themselves, but in their unification under a restoration-aligned generative formulation. Naively combining a generative flow model with a fidelity expert without our formulation leads to a fidelity collapse (33.69 → 27.60) as shown in Table~\ref{tbl:ablation_module}. Our task-aligned generative formulation plays a crucial role in unifying the fidelity and realism experts. Specifically, we reinterpret the generative vector field itself as a task-aligned residual error. This enables generative models to be optimized under restoration-oriented objectives, which mitigates a fidelity drop (33.69 → 33.05) as shown in Table~\ref{tbl:ablation_module}.
As a result, the PSNR improvement is not solely attributed to the fidelity expert, but our proposed task-aligned flow matching formulation also plays a crucial role in mitigating fidelity drop.
Furthermore, by jointly leveraging both the generative vector field and restoration-oriented vector field within a flow matching framework, our framework can bridge the gap between restoration and generation, allowing a single generative framework to operate as a deblurring model when blur images (or fidelity-expert results) are used as inputs and as a generative model when driven by noise.  

\begin{figure}[!tb]
  \centering
\begin{minipage}[!t]{.266\linewidth}
  \centering
  \centerline{\includegraphics[width=3.7cm]{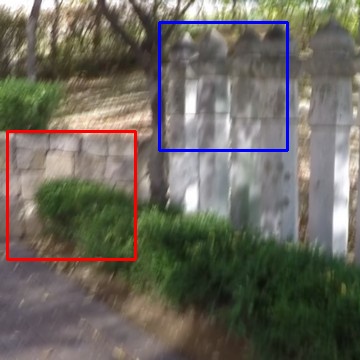}}
\end{minipage}
\begin{minipage}[!t]{.132\linewidth}
  \centering
  \centerline{\includegraphics[width=1.85cm]{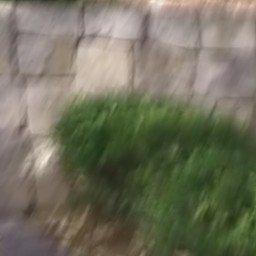}}
  \centerline{\includegraphics[width=1.85cm]{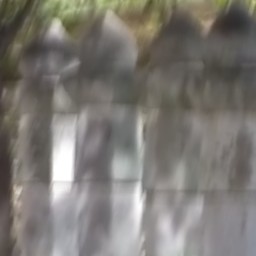}}
\end{minipage}
\begin{minipage}[!t]{.132\linewidth}
  \centering
  \centerline{\includegraphics[width=1.85cm]{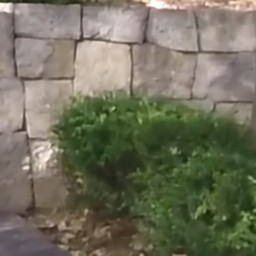}}
  \centerline{\includegraphics[width=1.85cm]{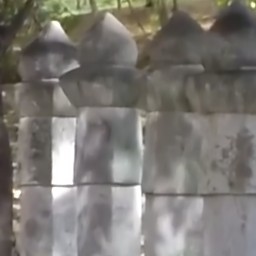}}
\end{minipage}
\begin{minipage}[!t]{.132\linewidth}
  \centering
  \centerline{\includegraphics[width=1.85cm]{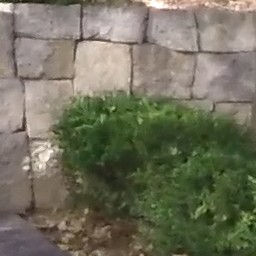}}
  \centerline{\includegraphics[width=1.85cm]{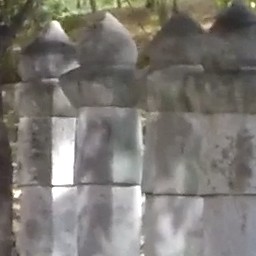}}
\end{minipage}
\begin{minipage}[!t]{.132\linewidth}
  \centering
  \centerline{\includegraphics[width=1.85cm]{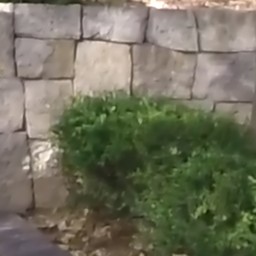}}
  \centerline{\includegraphics[width=1.85cm]{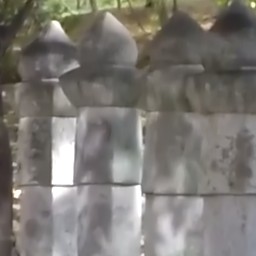}}
\end{minipage}
\begin{minipage}[!t]{.132\linewidth}
  \centering
  \centerline{\includegraphics[width=1.85cm]{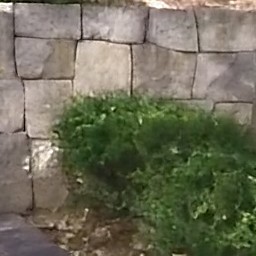}}
  \centerline{\includegraphics[width=1.85cm]{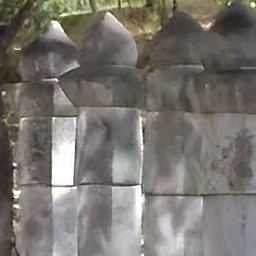}}
\end{minipage}
\vspace{-0.35cm}
\end{figure}
\begin{figure}[!tb]
  \centering
\begin{minipage}[!t]{.266\linewidth}
  \centering
  \centerline{\includegraphics[width=3.70cm]{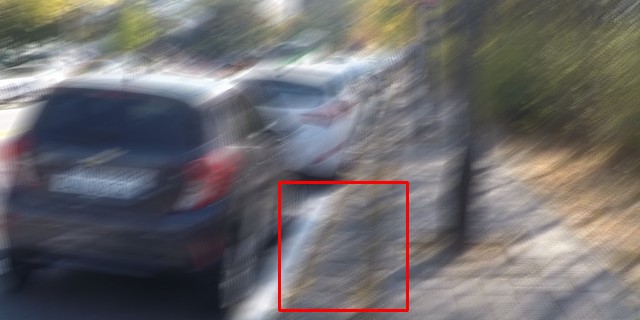}}
  \centerline{\includegraphics[width=3.70cm]{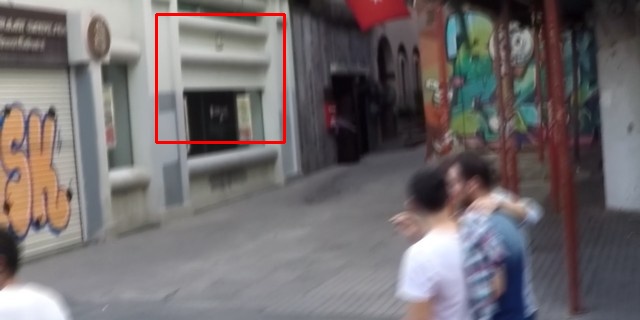}}
\end{minipage}
\begin{minipage}[!t]{.132\linewidth}
  \centering
  \centerline{\includegraphics[width=1.85cm]{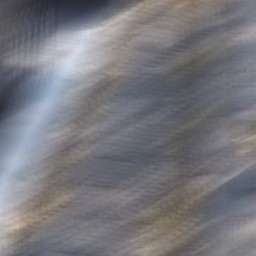}}
  \centerline{\includegraphics[width=1.85cm]{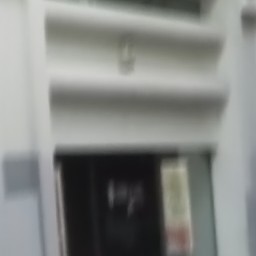}}
\end{minipage}
\begin{minipage}[!t]{.132\linewidth}
  \centering
  \centerline{\includegraphics[width=1.85cm]{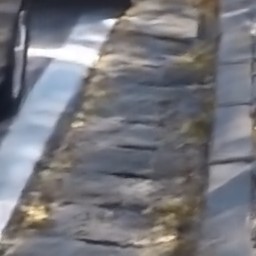}}
  \centerline{\includegraphics[width=1.85cm]{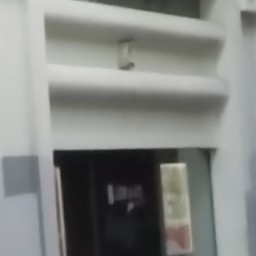}}
\end{minipage}
\begin{minipage}[!t]{.132\linewidth}
  \centering
  \centerline{\includegraphics[width=1.85cm]{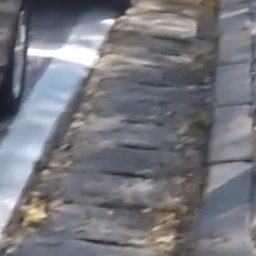}}
  \centerline{\includegraphics[width=1.85cm]{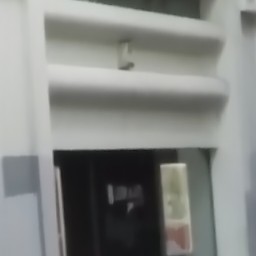}}
\end{minipage}
\begin{minipage}[!t]{.132\linewidth}
  \centering
  \centerline{\includegraphics[width=1.85cm]{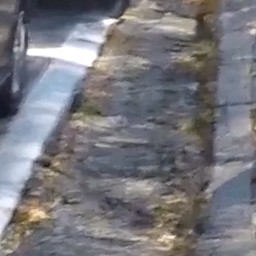}}
  \centerline{\includegraphics[width=1.85cm]{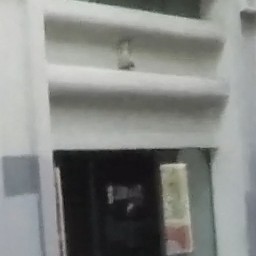}}
\end{minipage}
\begin{minipage}[!t]{.132\linewidth}
  \centering
  \centerline{\includegraphics[width=1.85cm]{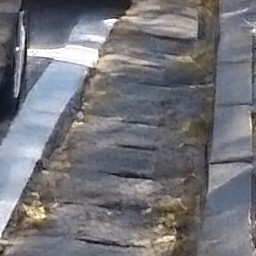}}
  \centerline{\includegraphics[width=1.85cm]{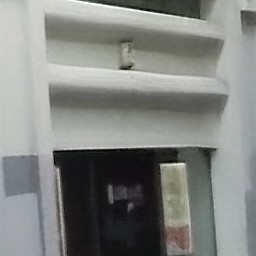}}
\end{minipage}
\vspace{-0.35cm}
\end{figure}
\begin{figure}[!t]
\begin{minipage}[!t]{.990\linewidth}
\centering\leftline{\;\;\;\;\;\;\;\;\;\;\;\;\;\;\;\;\;\;\;\;\;\;\;\;\;\;\;\;\;\;\;\;\;\;\;\;\;\;\;\;\;\;\;\;\;\;\;\;\small{Blur}\;\;\;\;\;\;\;\;\;\;\;\small{FFTFormer}\;\;\;\;\;\;\;\;\;\small{Hi-Diff}\;\;\;\;\;\;\;\;\;\;\small{IR-SDE}\;\;\;\;\;\small{DeblurFlow (ours)}}\medskip
\end{minipage}
\vspace{-0.3cm}
\caption{Qualitative comparison results on GoPro~\cite{gopro}. The proposed method produces realistic details and sharper results compared to restoration-based and diffusion-based methods.}
\vspace{-0.3cm}
\label{fig:main_visual_results}
\end{figure}
\setlength{\parindent}{0in}\textbf{Why our DeblurFlow path matters.} 
While the alternative path $\epsilon \rightarrow y-x$ can predict the residual error directly from noise, we intentionally adopt $y \rightarrow x$ for two reasons: (1) unlike noise-to-residual paths that require the model to hallucinate fidelity structure from scratch, using the blur image as the starting point provides a strong structural anchor, which is highly advantageous for restoration fidelity, and (2) despite starting from the blur image, the final target remain a clean image ($y \rightarrow x$) instead of a residual error ($\epsilon \rightarrow y-x$), which allows the model to fully exploit the powerful generative priors learned from large-scale clean images. Conversely, the $\epsilon \rightarrow y-x$ formulation shifts the target from a natural image to a residual error. This shift may break the natural image prior, leading to performance limitations. We present the empirical evidence in Table~\ref{tbl:eps_vs_yx}.

\section{Experiments} \label{sec:experiments}
\subsection{Experimental setup} \label{sec:experimental_setup}

\begin{table}[!t]
    \caption{Quantitative comparison on GoPro. Metrics are grouped into  \textbf{restoration fidelity} and \textbf{perceptual realism}. All methods are trained on the same dataset. We utilize 1-step sampling and NAFNet as the fidelity expert. The best results are indicated in bold.}
    \vspace{0.1cm}    
    \centering
    \scalebox{0.85}{
    \begin{tabular}{l|cccc|cccc}
        \hline
        \multirow{2}{*}{Methods} & \multicolumn{4}{c|}{\textbf{Restoration Fidelity}} 
        & \multicolumn{4}{c}{\textbf{Perceptual Realism}} \\
        \cline{2-5} \cline{6-9}
        & PSNR$\uparrow$ & SSIM$\uparrow$ & LPIPS$\downarrow$ & DISTS$\downarrow$
        & CLIPIQA$\uparrow$ & NIQE$\downarrow$ & MUSIQ$\uparrow$ & MANIQA$\uparrow$ \\
        \hline
        \multicolumn{9}{l}{\textbf{Restoration-Based Methods}}\\
        \hline
        MPRNet~\cite{mprnet}        & 32.66 & 0.9590 & 0.088 & 0.074 & 0.2537 & 5.16 & 44.18 & 0.5230 \\
        HINet~\cite{hinet}         & 32.77 & 0.9590 & 0.088 & 0.070 & 0.2513 & 5.08 & 44.07 & 0.5227 \\
        MIMOUNet+~\cite{mimounet}    & 32.44 & 0.9570 & 0.093 & 0.072 & 0.2505 & 5.04 & 43.76 & 0.5244 \\
        Restormer~\cite{restormer}    & 32.92 & 0.9610 & 0.084 & 0.072 & 0.2557 & 5.18 & 44.96 & 0.5269 \\
        Uformer~\cite{uformer}      & 32.97 & 0.9670 & 0.086 & 0.072 & 0.2588 & 5.18 & 44.81 & 0.5265 \\
        Stripformer~\cite{stripformer}  & 33.08 & 0.9620 & 0.077 & 0.068 & 0.2444 & 4.98 & 46.00 & 0.5345 \\
        NAFNet~\cite{nafnet}       & 33.69 & 0.9660 & 0.078 & 0.067 & 0.2603 & 5.10 & 45.28 & 0.5355 \\
        UFPNet~\cite{ufp}       & 34.06 & 0.9680 & 0.076 & 0.066 & 0.2573 & 5.11 & 45.33 & 0.5364 \\
        FFTFormer~\cite{fftformer}    & 34.21 & 0.9690 & 0.070 & 0.065 & \textbf{0.2649} & 4.97 & 46.14 & 0.5447 \\
        AdaRevD-L~\cite{adarev}    & \textbf{34.60} & \textbf{0.9720} & 0.071 & 0.067 & 0.2600 & 5.03 & 45.79 & 0.5440 \\
        \hline        
        \multicolumn{9}{l}{\textbf{Diffusion-Based Methods}}\\
        \hline
        IR-SDE~\cite{irsde}       & 30.70 & 0.9010 & \textbf{0.064} & - & - & - & - & - \\
        DiffBIR~\cite{diffbir}      & 26.15 & 0.8377 & 0.236 & 0.146 & - & - & - & - \\
        Hi-Diff~\cite{hidiff}      & 33.33 & 0.9640 & 0.079 & 0.071 & 0.2584 & 5.19 & 45.55 & 0.5360 \\
        DiffIR~\cite{diffir}       & 33.20 & 0.9630 & 0.078 & 0.070 & 0.2603 & 5.16 & 45.67 & 0.5388 \\
        FideDiff~\cite{fidediff}     & 28.79 & 0.9148 & 0.083 & 0.052 & 0.2121 & 4.42 & 45.57 & 0.5437 \\
        \textbf{DeblurFlow (ours)} 
                      & 33.05 & 0.9634 & \textbf{0.064} & \textbf{0.056}
                      & 0.2548 & \textbf{4.38} & \textbf{50.81} & \textbf{0.5668} \\
        \hline
    \end{tabular}
    }
    \label{tbl:gopro_merged}
\vspace{-0.3cm}
\end{table}
\setlength{\parindent}{0in}\textbf{Training and evaluation datasets.}
We train our DeblurFlow on GoPro~\cite{gopro}, and evaluate on GoPro, HIDE~\cite{hide}, RealBlur-J~\cite{realblur}, and RWBI~\cite{dbgan} to verify performance improvement and real-world generalization. GoPro contains $3,214$ blur–clean image pairs ($2,103$ for training and $1,111$ for evaluation). 
HIDE includes $2,025$ test paired images with diverse human-centric motions. RealBlur-J is a realistic dataset captured using a beam-splitter camera setup, consisting of $3,758$ training and $980$ test pairs in the sRGB domain. RWBI provides a large collection of real-world blurry images with spatially varying and complex blur patterns, consisting of $3,112$ test pairs. RWBI contains blur images without ground-truth clean counterparts. Hence, this RWBI allows evaluation only using no-reference metrics.

\setlength{\parindent}{0in}\textbf{Evaluation metrics.}
We use a comprehensive set of metrics to assess both restoration fidelity and perceptual realism. The fidelity metrics contain PSNR and SSIM~\cite{ssim} for distortion fidelity, and LPIPS~\cite{lpips} and DISTS~\cite{dists} for perceptual fidelity.
To evaluate perceptual realism, we adopt no-reference image quality metrics such as CLIP-IQA~\cite{clipiqa}, NIQE~\cite{niqe}, MUSIQ~\cite{musiq}, and MANIQA~\cite{maniqa}. CLIP-IQA estimates human perceptual quality based on multi-modal semantic alignment, while NIQE quantifies deviations from natural image statistics. MUSIQ and MANIQA evaluate aesthetic and perceptual realism through transformer- and attention-based network architectures, respectively. Finally, we compute the number of network parameters, Multiply–ACcumulate operations (MACs) based on $256 \times 256$ images, and runtime based on $2048 \times 2048$ images.

\setlength{\parindent}{0in}\textbf{Network architecture details.}
Our framework uses two complementary experts: a fidelity expert and a realism expert.
By default, we employ a pre-trained NAFNet~\cite{nafnet} as the fidelity expert.
The realism expert, i.e., our DeblurFlow, is built upon SANA-0.6B~\cite{sana}, which is one of the latent flow matching models.
For the LoRA implementation, we set the rank $r=32$ and the scaling factor $\alpha=64$. We apply LoRA to the query ($W_q$), key ($W_k$), value ($W_v$), and output ($W_o$) projection layers within the transformer blocks.
We adopt NAFNet-32~\cite{nafnet} for $r$-space encoder-decoder.
We present more details on network architecture and implementation in Section~\ref{sec:app_details}.

\subsection{Fidelity-Realism Motion Deblurring} \label{sec:result_gopro}
Diffusion-based deblurring models such as IR-SDE~\cite{irsde} and DiffBIR~\cite{diffbir} inherently struggle to produce high-fidelity results, as their generation-oriented objectives are designed to synthesize images from noise rather than reconstructing residual errors, as shown in Table~\ref{tbl:gopro_merged}. Recently, FideDiff~\cite{fidediff} introduces a Kernel ControlNet to estimate and use blur kernels, improving structural fidelity. However, it still lags behind restoration-based deblurring models in terms of reconstruction fidelity. Specifically, IR-SDE, DiffBIR, and FideDiff achieve 30.70 dB, 26.15 dB, and 28.79 dB in PSNR, respectively.
To address the fidelity collapse commonly observed in generative deblurring models, we assign complementary roles to a fidelity expert and a fidelity-preserving realism expert. 
This decomposition effectively mitigates fidelity collapse, achieving 33.05 dB in PSNR. At the same time, our DeblurFlow attains superior perceptual scores compared to diffusion-based methods across NIQE, MUSIQ, and MANIQA metrics, as shown in Table~\ref{tbl:gopro_merged}.
Furthermore, the diffusion-assisted deblurring methods like Hi-Diff~\cite{hidiff} and DiffIR~\cite{diffir} adopt restoration-based networks and leverage the diffusion model only as a supportive cue. Consequently, they struggle to fully exploit the powerful generative priors of pre-trained diffusion models for enhancing perceptual realism. While they achieve high-fidelity results, their ability to enhance perceptual realism remains limited, as shown in Table~\ref{tbl:gopro_merged}.

\begin{table*}[!t]
    \caption{Performance comparison on \textbf{HIDE}~\cite{hide} and \textbf{RealBlur-J}~\cite{realblur}. All methods are trained on the same dataset. We utilize 1-step sampling and NAFNet~\cite{nafnet} as the fidelity expert. The best results are indicated in bold.}
    \vspace{-0.1cm}
    \centering
    \setlength{\tabcolsep}{5pt}
    \scalebox{0.78}{
    \begin{tabular}{l|ccccc|ccccc}
        \hline
        \multirow{2}{*}{Methods} & \multicolumn{5}{c|}{\textbf{HIDE}} 
        & \multicolumn{5}{c}{\textbf{RealBlur-J}} \\
        \cline{2-6} \cline{7-11}
        & PSNR$\uparrow$ & LPIPS$\downarrow$ & DISTS$\downarrow$ & NIQE$\downarrow$ & MUSIQ$\uparrow$
        & PSNR$\uparrow$ & LPIPS$\downarrow$ & DISTS$\downarrow$ & NIQE$\downarrow$ & MUSIQ$\uparrow$ \\
        \hline
        MIMOUNet+~\cite{mimounet}    & 29.99 & 0.125 & 0.075 & 4.44 & 52.10 & 27.62 & 0.196 & 0.133 & 5.54 & 43.20 \\
        Restormer~\cite{restormer}   & 31.22 & 0.109 & 0.073 & 4.66 & 53.63 & 28.96 & 0.156 & 0.116 & 5.23 & 48.43\\
        Stripformer~\cite{stripformer} & 31.03 & 0.105 & 0.067 & 4.35 & 54.74 & - & 0.154 & 0.114 & 5.09 & 47.09\\
        NAFNet~\cite{nafnet}         & 31.33 & 0.104 & 0.067 & 4.32 & 54.34 & 28.32 & 0.168 & 0.119 & 5.05 & 47.07\\
        FFTFormer~\cite{fftformer}   & \textbf{31.62} & {0.096} & {0.065} & 4.36 & 55.21 & 27.73 & 0.184 & 0.123 & 5.38 & 46.60 \\
        UFPNet~\cite{ufp}            & - & \textbf{0.093} & 0.068 & 4.35 & 54.74 & 29.87 & 0.144 & 0.110 & 5.45 & 51.31\\
        Hi-Diff~\cite{hidiff}        & 31.41 & 0.105 & 0.073 & 4.60 & 54.61 & {28.90} & 0.147 & 0.108 & 5.24 & 50.55  \\
        DiffIR~\cite{diffir}         & 31.38 & 0.101 & 0.071 & 4.58 & 54.58 & 28.80 & 0.154 & 0.109 & 5.20 & 48.79\\
        DiffBIR~\cite{diffbir}       & - & 0.209 & 0.124 & - & - & - & 0.258 & 0.159 & - & -\\
        \textbf{DeblurFlow (ours)} 
                      & 30.90 & {0.096} & \textbf{0.054} & \textbf{3.66} & \textbf{59.88} 
                      & \textbf{30.60} & \textbf{0.114} & \textbf{0.095} & \textbf{4.71} & \textbf{54.96} \\
        \hline
    \end{tabular}
    }
    \label{tbl:hide_realblur_merged}
    \vspace{-0.2cm}
\end{table*}

\subsection{Comprehensive Evaluations} \label{sec:several_eval}
\setlength{\parindent}{0in}\textbf{Results on HIDE and RealBlur-J.} 
To evaluate our method on different datasets, we report quantitative comparisons on both HIDE and RealBlur-J. As shown in Table~\ref{tbl:hide_realblur_merged}, for HIDE, our DeblurFlow significantly improves perceptual fidelity (e.g., LPIPS and DISTS) and perceptual realism (e.g., NIQE and MUSIQ) compared to its counterpart, NAFNet~\cite{nafnet}. Specifically, our DeblurFlow exhibits a performance improvement from 0.104 to 0.096 dB in LPIPS, and 54.34 to 59.88 in MUSIQ.
Similar improvements are observed on the RealBlur-J, where DeblurFlow enhances perceptual realism while simultaneously improving restoration fidelity.
These improvements originate from the generative priors of the flow matching models trained with large-scale natural clean images, which enable our DeblurFlow to generalize toward more perceptually realistic image reconstructions.
Note that our DeblurFlow achieves comparable LPIPS than UFPNet (0.093 vs. 0.096), with 4$\times$ higher efficiency as shown in Table~\ref{tbl:gmacs}.
We present the visual results on RealBlur-J as shown in Fig.~\ref{fig:app_result_realblur}.

\begin{table*}[!t]
\centering
\begin{minipage}[t]{0.48\textwidth}
    \captionof{table}{Performance comparison on \textbf{RWBI}~\cite{dbgan}. All methods are trained on the same dataset. DeblurFlow and DeblurFlow-3S correspond to 1-step and 3-step sampling.}
    \centering
    \scalebox{0.72}{
    \setlength{\tabcolsep}{1.0pt}
    \begin{tabular}{lcccc}
        \toprule
        Methods & CLIPIQA$\uparrow$ & NIQE$\downarrow$ & MUSIQ$\uparrow$ & MANIQA$\uparrow$\\
        \midrule
        MIMOUNet+ & 0.2330 & 5.12 & 43.90 & 0.4708 \\
        Restormer & 0.2458 & 5.23 & 43.79 & 0.5053 \\
        FFTFormer & 0.2319 & 4.95 & 43.01 & 0.5029 \\
        Hi-Diff & 0.2589 & 5.29 & 47.78 & 0.5122 \\
        DiffIR & 0.2435 & 5.37 & 44.40 & 0.4954 \\
        NAFNet & 0.2384 & 4.95 & 46.57 & 0.4894 \\
        \textbf{DeblurFlow (ours)} & {0.2867} & {4.43} & {50.64} & {0.5091} \\
        \textbf{DeblurFlow-3S (ours)} & \textbf{0.3078} & \textbf{4.24} & \textbf{51.96} & \textbf{0.5143} \\
        \bottomrule
    \end{tabular}
    }
    \label{tbl:rwbi}
\end{minipage}\hfill
\begin{minipage}[t]{0.48\textwidth}
\captionof{table}{Computational cost and deblurring performance. The computational cost of our DeblurFlow includes all components such as fidelity expert (NAFNet), r-space encoder-decoder and flow model. All methods are trained on the same dataset. DeblurFlow and DeblurFlow-3S correspond to 1-step and 3-step sampling.}
\centering
\scalebox{0.87}{
\setlength{\tabcolsep}{1.0pt}
\footnotesize
\begin{tabular}{l|ccc|c}
\hline
Methods 
& \# Params & MACs & Runtime 
& LPIPS\;MUSIQ\\
\hline
MIMOUNet+  & 16.1M & 154.4G & 0.36s & 0.093\;\;43.76 \\
NAFNet    & 67.7M &  63.6G & 0.26s & 0.078\;\;45.28 \\
Restormer & 26.0M & 141.0G & 1.68s & 0.084\;\;44.96 \\
FFTFormer & 14.8M & 131.4G & 2.66s & 0.070\;\;46.14 \\
UFPNet & 79.7M & 242.8G & 1.65s & 0.076\;\;46.14 \\
Hi-Diff   & 28.4M & 142.6G & 1.54s & 0.079\;\;45.55 \\
DiffIR    & 28.1M & 112.7G & 1.40s & 0.078\;\;45.67 \\
\textbf{DeblurFlow (ours)}  & 768.5M & 115.4G & 0.41s & \textbf{0.064}\;\;{50.81} \\
\textbf{DeblurFlow-3S (ours)}  & 768.5M & 219.1G & 0.71s & {0.078}\;\;\textbf{53.29} \\
\hline
\end{tabular}
}
\label{tbl:gmacs}
\end{minipage}
\vspace{-0.3cm}
\end{table*}

\setlength{\parindent}{0in}\textbf{Results on RWBI.} 
To assess the real-world robustness of our method, we evaluate DeblurFlow on the RWBI~\cite{dbgan}, which comprises blur images collected from real-world scenarios.
As shown in Table~\ref{tbl:rwbi}, our DeblurFlow consistently outperforms both restoration-based methods~\cite{mimounet,restormer,fftformer} and diffusion-based methods~\cite{diffir,hidiff} across all no-reference perceptual quality metrics. In particular, our DeblurFlow achieves CLIPIQA (0.2867), NIQE (4.43), MUSIQ (50.64) scores, and DeblurFlow-3S achieves MANIQA (0.5143) scores, indicating superior perceptual quality. 
Notably, our DeblurFlow-3S outperforms Hi-Diff in MANIQA (0.5143 vs. 0.5122) while remaining more efficient. Specifically, our DeblurFlow-3S requires only 0.71s with 3-step sampling, whereas Hi-Diff~\cite{hidiff} takes 1.54s, as shown in Table~\ref{tbl:gmacs}.
These results demonstrate that our DeblurFlow, built upon the natural clean image priors of pre-trained flow matching models, generalizes effectively to real-world blur scenarios. 
We present the visual results on RWBI as shown in Fig.~\ref{fig:app_result_rwbi}.

\begin{table*}[!t]
\centering
\begin{minipage}[t]{0.48\textwidth}
\captionof{table}{Computational cost comparison between DeblurFlow-VAE and DeblurFlow with different sampling steps.}
\centering
\scalebox{0.82}{
\begin{tabular}{lccc}
\hline
Methods & \# Steps & GMACs & Runtime (s) \\
\hline
DeblurFlow-VAE & 1  & 518.64 & 0.79 \\
               & 5  & 2338.64 & 2.91 \\
               & 10 & 4613.64 & 5.56 \\
\hline
DeblurFlow (ours) & 1  & 115.49 & 0.41 \\
                  & 5  & 322.85 & 1.01 \\
                  & 10 & 582.05 & 1.76 \\
\hline
\end{tabular}
}
\label{tab:efficiency_comparison}
\end{minipage}\hfill
\begin{minipage}[t]{0.48\textwidth}
\captionof{table}{Ablation study on our components. Each component contributes to improving performance. We use NAFNet as the fidelity expert. We utilize 1-step sampling.}
\centering
\scalebox{0.83}{
\begin{tabular}{c|c|c|c|c}
\hline{Fidelity Expert}             & {\checkmark}   & {\checkmark} & {\checkmark}      & {\checkmark}\\
{Flow Model}             & {}      & {\checkmark} & {\checkmark}      & {\checkmark}\\
{Residual Loss}                     & {}      & {}          & {\checkmark}      & {\checkmark}\\
{r-Space Latents}              & {}     & {}          & {}      & {\checkmark}\\
\hline{PSNR $\uparrow$}    & {$\textbf{33.69}$} & {$27.60$}   & {$32.40$}           & {${33.05}$}\\
{LPIPS} $\downarrow$       & {$0.078$} & {$0.120$}   & {$0.069$}         & {$\textbf{0.064}$}  \\
{MUSIQ} $\uparrow$         & {$45.28$} & {$48.96$}   & {${47.16}$}  & {$\textbf{50.81}$}\\
{MANIQA} $\uparrow$        & {$0.535$} & {$0.556$}   & {$0.559$}         & {$\textbf{0.566}$}  \\
\hline{GMACs}              & {$63.64$} & {$523.83$}  & {$523.83$}         & {${115.49}$}\\
\hline
\end{tabular}
}
\label{tbl:ablation_module}
\end{minipage}
\vspace{-0.3cm}
\end{table*}

\subsection{Ablation Study} \label{sec:ablation}
\setlength{\parindent}{0in}\textbf{Computational costs.}
{
We comprehensively analyze the computational costs, including model parameters, GMACs, and runtime in Table~\ref{tbl:gmacs}. To ensure a fair comparison with previously reported GMACs of existing deblurring methods, we compute it on $256 \times 256$ images.
We measure the runtime on high-resolution $2048 \times 2048$ images using a single NVIDIA H100 GPU.
A key advantage of our DeblurFlow is its practical inference speed.
As shown in Table~\ref{tbl:gmacs}, our DeblurFlow requires only 0.41s to process a 2K resolution image. This is approximately $3.4\times$ faster than DiffIR and $6.5\times$ faster than FFTFormer.
Despite its high inference efficiency, our DeblurFlow delivers a significant improvement in perceptual fidelity (LPIPS) and realism (MUSIQ).
These results demonstrate that our DeblurFlow overcomes the computational bottlenecks of existing generative deblurring models, making it practical for real-world applications.}

\setlength{\parindent}{0in}\textbf{Latent space efficiency (r-space vs. VAE space).}
{
Table~\ref{tab:efficiency_comparison} highlights the efficiency of our r-space over the VAE latent space. While the computational cost of VAE latents increases steeply as the number of sampling steps grows, our r-space remains computationally efficient, validating the suitability of r-space latents for iterative sampling.
}

\setlength{\parindent}{0in}\textbf{Contribution by each module.}
{
To verify the impact of each proposed module, we progressively integrate our three components (Flow Model, Residual Loss, and r-Space Latent into the fidelity expert baseline. 
As shown in Table~\ref{tbl:ablation_module}, incorporating the generative flow model improves perceptual realism (e.g., MUSIQ 45.28 $\rightarrow$ 48.96) by leveraging its inherent generative priors.
However, it suffers a significant fidelity drop (PSNR 33.69 $\rightarrow$ 27.60 dB). This clearly confirms that the standard generation objectives fail to preserve the initial fidelity.
Introducing the Residual Loss aligns the objective with deblurring tasks, effectively mitigating this drop and boosting PSNR to 32.40 dB while enhancing perceptual fidelity (LPIPS 0.069).
Finally, substituting the standard VAE latent space with the r-Space further strengthens both fidelity and realism (e.g., PSNR 33.05, MUSIQ 50.81, MANIQA 0.566) while reducing computational cost by nearly 5×. This confirms that the r-space latents remove the fidelity bottleneck imposed by generation-oriented VAE latents, leading to higher reconstruction fidelity. Moreover, the r-space is explicitly tailored for image residual decoding, whereas conventional VAEs are optimized for reconstructing clean images, which translates into more accurate residual recovery and improved perceptual realism.
For further insights, we perform more ablation studies in Section~\ref{sec:app_ablation}.}

\begin{wraptable}{r}{0.46\linewidth}
\vspace{-0.8em}
\centering
\caption{Comparison among several paths.}
\label{tbl:eps_vs_yx}
\setlength{\tabcolsep}{4pt}
\scalebox{0.84}{
\begin{tabular}{lccc}
\toprule
Method & PSNR$\uparrow$ & LPIPS$\downarrow$ & MUSIQ$\uparrow$ \\
\midrule
$\epsilon \rightarrow x$ & 27.60 & 0.120 & 48.96 \\
$\epsilon \rightarrow y-x$ & 32.52 & 0.089 & 44.23 \\
$y \rightarrow x$ (ours) & \textbf{33.05} & \textbf{0.064} & \textbf{50.81} \\
\bottomrule
\end{tabular}}
\vspace{-1.2em}
\end{wraptable}
\setlength{\parindent}{0in}\textbf{Blur-to-clean vs. noise-to-residual.}
To empirically validate our blur-to-clean design choice, we compare our direct mapping $y \rightarrow x$ with the alternative formulation $\epsilon \rightarrow y-x$. As shown in Table~\ref{tbl:eps_vs_yx}, both formulations benefit from the task-aligned residual loss and achieve reasonably high PSNR. However, the direct mapping used in DeblurFlow yields stronger fidelity (33.05 dB vs. 32.52 dB) and substantially better perceptual quality (LPIPS 0.064 and MUSIQ 50.81). In contrast, the $\epsilon \rightarrow y-x$ formulation exhibits clear limitations in effectively leveraging pretrained generative priors, leading to noticeably weaker realism (MUSIQ 44.23).


\vspace{-0.1cm}
\section{Conclusions} \label{sec:conclusion}
We present DeblurFlow, a restoration-aligned generative deblurring framework that repurposes pretrained flow models for blur-to-clean transport. By reformulating the vector field as a task-aligned residual error and pairing it with r-space and dual-expert sampling, our DeblurFlow achieves high restoration fidelity and perceptual realism while remaining computationally practical. Our method can be extended to other restoration tasks such as image denoising~\cite{key_denoising} and image super-resolution~\cite{key_sr}. Crucially, our DeblurFlow functionally aligns with standard generative flow models, suggesting that restoration and generation need not be viewed as distinct tasks, but can be explored within a single generative flow model.

\bibliographystyle{unsrtnat}
\bibliography{neurips_2026}

\newpage
\appendix
\onecolumn
\section{Broader Impacts and Limitations}\label{sec:broader_impact}
\setlength{\parindent}{0in}\textbf{Broader Impacts.} 
Our method offers several potential societal benefits, including enhanced visual reliability for autonomous driving, robotics, and mobile photography, where motion blur can degrade performance.
However, we recognize the ethical considerations inherent in generative restoration. Like all generative models, there is a risk of hallucinating plausible but inaccurate textures, which means the outputs should not be treated as ground-truth evidence in safety-critical or medical applications. We mitigate these risks by prioritizing high-fidelity reconstruction constraints in both the training and sampling process.

\setlength{\parindent}{0in}\textbf{Limitations.} While we employ dual experts to better balance fidelity and realism, relying on an external fidelity expert remains a limitation of the current framework. A key direction for future work is to eliminate the need for an external fidelity expert by internalizing it within a single generative deblurring model.

\section{Computing Resources}\label{sec:computing_resource}
Our experiments are conducted on an internal cluster using an Intel Xeon Gold 6448Y CPU (24 cores) and four GPUs, each equipped with 80 GB of memory. Training our primary model (768M parameters) on the GoPro dataset required 28 hours for 1000 epochs, while inference for the full test set took 0.63 hours, resulting in a total of 28.63 compute hours for the reported results. The overall research project required additional computational costs beyond these figures to account for preliminary experiments, debugging, and various ablation studies conducted during the development phase.

\section{Implementation Details}\label{sec:app_details}
\setlength{\parindent}{0in}\textbf{Network architecture details.}
Our framework consists of two complementary experts: a fidelity expert and a realism expert.
By default, we employ a pre-trained NAFNet~\cite{nafnet} as the fidelity expert.
The realism expert, i.e., our DeblurFlow, is built upon SANA-0.6B~\cite{sana}, which is one of the latent flow matching models.
We adopt NAFNet-32 with encoder blocks $\{2,2,2,2,18\}$ and a middle block $\{4\}$ for the $r$-space encoder, and decoder blocks $\{2,2,2,2,2\}$ for the $r$-space decoder.
Note that NAFNet-32 denotes the NAFNet architecture configured with a base channel width of 32.
The $r$-space encoder is equipped with a projection layer (32$\times$2$\times$2$\times$2$\times$2$\times$2 $\rightarrow$ 32 channel), where the output 32 channels serve as the base input channels of the flow matching model, and the $r$-space decoder comes with a projection layer (32 $\rightarrow$ 32$\times$2$\times$2$\times$2$\times$2$\times$2 channel), to ensure compatibility with the pre-trained flow matching model.
For the LoRA implementation, we set the rank $r=32$ and the scaling factor $\alpha=64$. We apply LoRA to the query ($W_q$), key ($W_k$), value ($W_v$), and output ($W_o$) projection layers within the transformer blocks.

\vspace{0.02in}
\setlength{\parindent}{0in}\textbf{Implementation details.} 
We train our DeblurFlow on GoPro~\cite{gopro} with random crops of size $512 \times 512$. The model is trained for $1{,}000$ epochs with a batch size of $32$ using $4$ NVIDIA H100 GPUs. We adopt the AdamW~\cite{adamw} optimizer with $(\beta_1, \beta_2) = (0.9, 0.9)$ and a weight decay of $1e^{-3}$. The learning rate follows a cosine annealing schedule starting from $1e^{-4}$.
During training, we sample the full degradation pairs ($y \to x$) and the expert-dependent pairs ($f_\psi(y) \to x$) with probabilities of 0.7 and 0.3, respectively.
All ablation studies are conducted on GoPro dataset.

\section{Additional Ablation Studies}\label{sec:app_ablation}
\subsection{Co-Training Strategy}\label{sec:app_discussion_using_fidelity_expert}
\begin{figure}[!t]
\begin{minipage}[!ht]{.32\linewidth}
  \centering
  \centerline{\includegraphics[width=4.5cm]{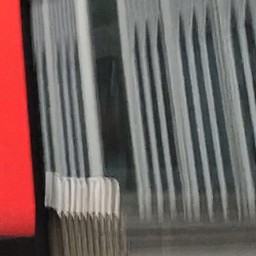}}
  \centerline{\small{(d) IR-SDE~\cite{irsde}}}\medskip
\end{minipage}
\begin{minipage}[!ht]{.32\linewidth}
  \centering
  \centerline{\includegraphics[width=4.5cm]{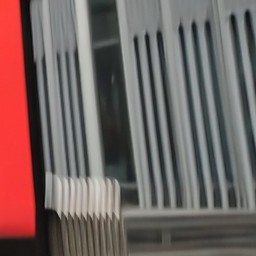}}
  \centerline{\small{(e) DiffIR~\cite{diffir}}}\medskip
\end{minipage}
\begin{minipage}[!ht]{.32\linewidth}
  \centering
  \centerline{\includegraphics[width=4.5cm]{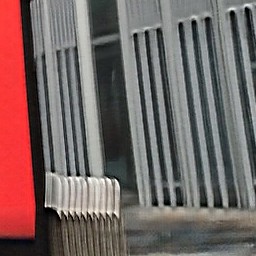}}
  \centerline{\small{(f) DeblurFlow (ours)}}\medskip
\end{minipage}
\vspace{-0.3cm}
\caption{Visual comparison results on RWBI~\cite{dbgan}.}
\vspace{-0.3cm}
\label{fig:app_result_rwbi}
\end{figure}

{
A critical design choice in our framework is to adopt a co-training strategy. 
Given that our fidelity-realism sampling originates from the fidelity expert's estimate ($f_\psi(y)$), a potential distribution gap arises if the model is trained solely on blur-to-clean mapping ($y \to x$). To mitigate this training-sampling discrepancy, we first adopt the expert-dependent mapping ($f_\psi(y) \to x$) for the training objective. However, relying exclusively on this mapping forces the model to overfit to the specific artifacts of the fidelity expert, limiting its generalization capability. To address this, we extend the training objective beyond the expert-dependent mapping by incorporating the full degradation mapping ($y \to x$).
This dual training ensures domain alignment while maintaining the model's capability to prevent overfitting to specific fidelity experts and enable robustness to challenging motion blur beyond the coverage of fidelity experts, as shown in Fig.~\ref{fig:app_result_rwbi}.
As reported in Table~\ref{tab:ablation_input_type}, the results show a clear perception-distortion trade-off governed by the input ratio. A higher reliance on the fidelity expert’s estimate $f_{\psi}(y)$ (e.g., 0:100) yields superior restoration fidelity (highest PSNR) because $f_{\psi}(y)$ provides a structurally accurate starting point, so that the model excels at minimizing distortion relative to the ground truth. Conversely, training on the full degradation trajectory $y \to x$ (e.g., 100:0) maximizes perceptual realism (highest MUSIQ/MANIQA), as the generative flow model is forced to synthesize high-frequency textures to recover the underlying clean images from blur observations. To strike an optimal balance, we choose a 70:30 ratio for all experiments. This configuration provides high perceptual quality while maintaining competitive fidelity.}

 \begin{table}[ht]
    \caption{Ablation study on co-training data ratio. All methods are trained and evaluated on the same dataset. We utilize 1-step sampling for our DeblurFlow. Ours is highlighted in bold.}
    \vspace{0.1cm}    
    \centering
    \scalebox{0.85} {
    \begin{tabular}{ccccccc}
        \toprule
        \makecell{Co-Training Data Ratio \\ ($y$ : $f_{\psi}(y)$)} & PSNR $\uparrow$ & LPIPS $\downarrow$ & CLIPIQA $\uparrow$ & NIQE $\downarrow$ & MUSIQ $\uparrow$ & MANIQA $\uparrow$ \\
        \midrule
        0 : 100 & 33.54 & 0.064 & 0.2503 & 4.57 & 46.55 & 0.5363 \\
        30 : 70 & 33.38 & 0.062 & 0.2531 & 4.45 & 48.93 & 0.5532 \\
        50 : 50 & 33.24 & 0.063 & 0.2547 & 4.44 & 49.86 & 0.5615 \\
        \textbf{70 : 30} & \textbf{33.05} & \textbf{0.064} & \textbf{0.2548} & \textbf{4.38} & \textbf{50.81} & \textbf{0.5668} \\
        100 : 0 & 32.39 & 0.071 & 0.2532 & 4.37 & 52.56 & 0.5733 \\
        \bottomrule
    \end{tabular} }
    \label{tab:ablation_input_type}
\end{table}

\subsection{Effects on Fidelity Experts}\label{sec:app_fidelity_expert}
{Leveraging a stronger fidelity expert provides a better initialization, which directly contributes to improved deblurring performance. Since our DeblurFlow adopts co-training strategy as discussed in Section~\ref{sec:app_discussion_using_fidelity_expert}, it can be compatible to various fidelity experts.
To confirm this, we experiment with several fidelity experts, including MIMO-UNet+~\cite{mimounet}, Restormer~\cite{restormer}, NAFNet~\cite{nafnet}, and FFTFormer~\cite{fftformer}.}
As summarized in Table~\ref{tab:network_performance}, our DeblurFlow consistently improves performance across all fidelity experts, confirming that our method is not specialized for a certain fidelity expert but generalizes well to diverse deblurring-based models.

\begin{table}[ht]
    \caption{Performance comparison of different fidelity networks. All methods are trained and evaluated on the same dataset. We utilize 1-step sampling for our DeblurFlow.}
    \vspace{0.1cm}
    \centering
    \setlength{\tabcolsep}{5pt} 
    \scalebox{0.95} {
    \begin{tabular}{lcccc}
        \toprule
        Network & PSNR $\uparrow$ & LPIPS $\downarrow$ & NIQE $\downarrow$ & MUSIQ $\uparrow$ \\
        \midrule
        MIMOUNet+~\cite{mimounet} & \textbf{32.44} & 0.093 & 5.04 & 43.76 \\
        + DeblurFlow & 31.84 & \textbf{0.075} & \textbf{4.39} & \textbf{49.33} \\
        \hline Restormer~\cite{restormer} & \textbf{32.92} & 0.084 & 5.18 & 44.96 \\
        + DeblurFlow & 32.29 & \textbf{0.071} & \textbf{4.48} & \textbf{50.79} \\
        \hline NAFNet~\cite{nafnet} & \textbf{33.69} & 0.078 & 5.10 & 45.28 \\
        + DeblurFlow & 33.05 & \textbf{0.064} & \textbf{4.38} & \textbf{50.81} \\
        \hline {FFTFormer~\cite{fftformer}} & \textbf{34.21} & 0.070 & 4.97 & 46.14 \\
        + DeblurFlow & 33.48 & \textbf{0.059} & \textbf{4.30} & \textbf{52.18} \\
        \bottomrule
    \end{tabular} }
    \label{tab:network_performance}
\end{table}

\subsection{Effects on the Number of Sampling Steps}\label{sec:app_n_steps}
\begin{figure}[ht]
\begin{minipage}[!t]{.24\linewidth}
  \centering
  \centerline{\includegraphics[width=2.04cm]{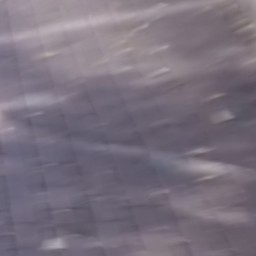}}
  \centerline{\small{(a)}}\medskip
\end{minipage}
\begin{minipage}[!t]{.24\linewidth}
  \centering
  \centerline{\includegraphics[width=2.04cm]{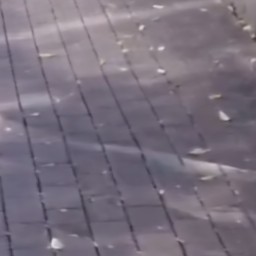}}
  \centerline{\small{(b)}}\medskip
\end{minipage}
\begin{minipage}[!t]{.24\linewidth}
  \centering
  \centerline{\includegraphics[width=2.04cm]{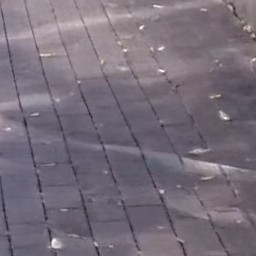}}
  \centerline{\small{(c)}}\medskip
\end{minipage}
\begin{minipage}[!t]{.24\linewidth}
  \centering
  \centerline{\includegraphics[width=2.04cm]{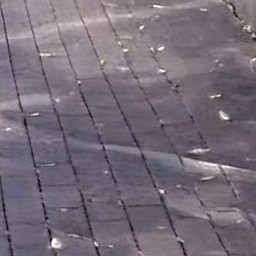}}
  \centerline{\small{(d)}}\medskip
\end{minipage}
\vspace{-0.3cm}
\caption{Effect of expert types and sampling steps: (a) Blur input, (b) Fidelity expert only, (c) Fidelity-realism sampling (1-step), and (d) Fidelity-realism sampling (3-step).}
\vspace{-0.3cm}
\label{fig:qual_results_sampling_steps}
\end{figure}

\begin{figure}[ht]
\begin{minipage}[!t]{.48\linewidth}
  \centering
  \centerline{\includegraphics[width=6.2cm]{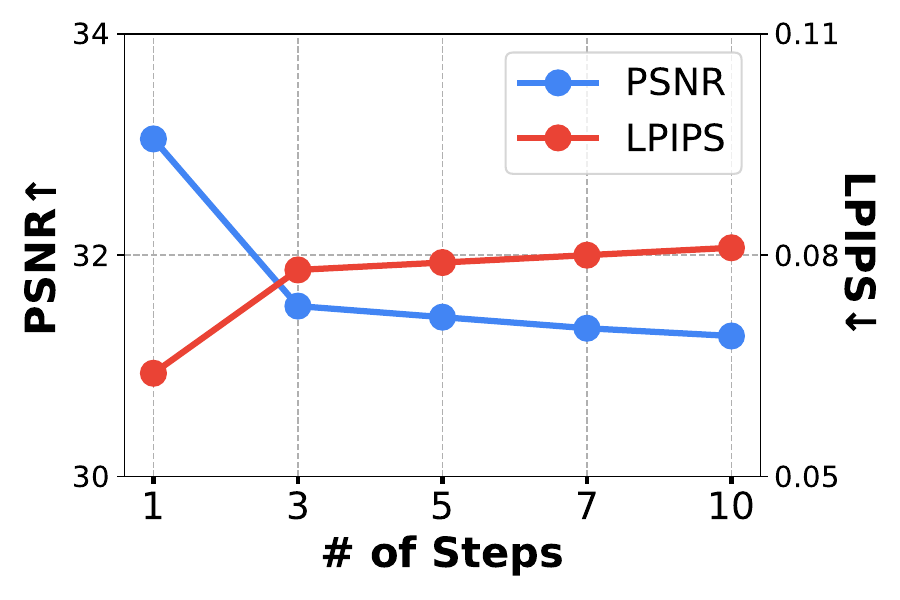}}
  \centerline{\small{(a)}}\medskip
\end{minipage}
\begin{minipage}[!t]{.48\linewidth}
  \centering
  \centerline{\includegraphics[width=6.2cm]{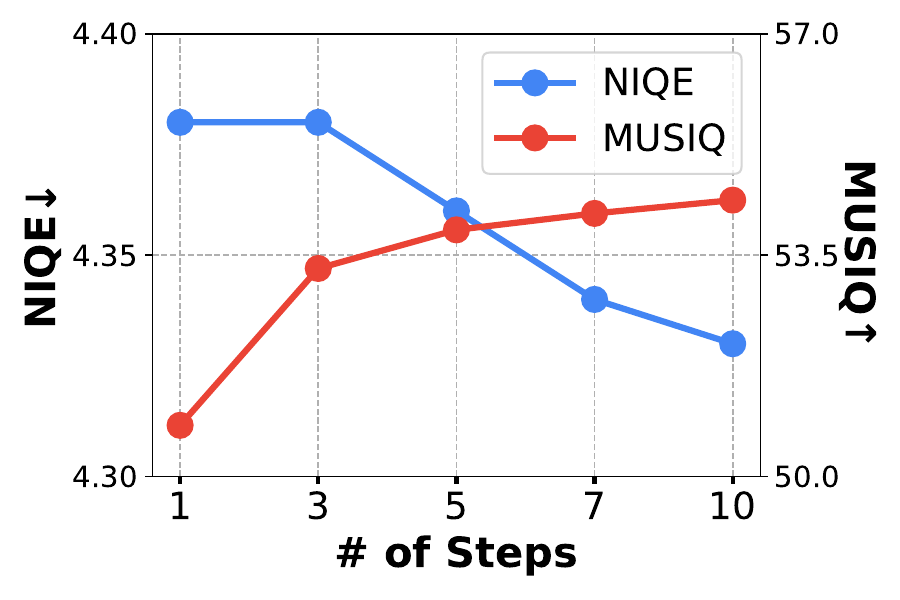}}
  \centerline{\small{(b)}}\medskip
\end{minipage}
\vspace{-0.3cm}
\caption{Ablation study on \# of sampling steps. (a) restoration fidelity (PSNR, LPIPS) and (b) perceptual realism (NIQE, MUSIQ).}
\label{fig:ablation_timesteps}
\end{figure}

We analyze the performance variation according to the number of steps in the sampling process.
As observed in Fig.~\ref{fig:ablation_timesteps}, the fidelity metrics such as PSNR and LPIPS remain relatively stable even as the number of sampling steps increases, since our fidelity-preserving residual learning mitigates this degradation by aligning the training objective with the deblurring task.
Furthermore, increasing the number of sampling steps from 1 to 20 leads to a gradual enhancement in perceptual realism like NIQE and MUSIQ scores, and increasing the number of steps yields more realistic visual results as shown in Fig.~\ref{fig:qual_results_sampling_steps}.
Notably, we observe that our DeblurFlow attains strong restoration fidelity and perceptual realism with 1-step sampling. This result suggests that our restoration-aligned formulation and dual-expert strategy substantially reduces the need for expensive multi-step sampling.

\subsection{Effects on Skip Connections}\label{sec:app_skip_conn}
To address the effect of skip connections, we conduct an ablation study by removing them. As shown in Table~\ref{tab:skip_connection_ablation}, the absence of skip connections leads to a clear decrease in restoration fidelity (33.05 $\rightarrow$ 30.52). This is because the decoder no longer receives the structural anchor and low-frequency content directly from the blur observation (or high-fidelity initialization). However, without skip connections, our model (30.52 dB) still outperforms the standard flow-based deblurring baseline (27.60 dB), demonstrating that our task-aligned residual formulation provides fidelity gains regardless of the skip connection design. Furthermore, the perceptual realism metrics (e.g., MUSIQ, MANIQA) improve when skip connections are removed (MUSIQ 50.81 $\rightarrow$ 53.98), indicating that the flow model contributes more to high-frequency texture synthesis in their absence. Nevertheless, since our goal is to achieve high fidelity and realism simultaneously, we adopt skip connections as a deliberate design choice.

\begin{table}[ht]
    \caption{Ablation study on skip connections.}
    \vspace{0.1cm}
    \centering
    \setlength{\tabcolsep}{6pt}
    \scalebox{0.92}{
    \begin{tabular}{lcccc}
        \toprule
        Method & PSNR$\uparrow$ & SSIM$\uparrow$ & MUSIQ$\uparrow$ & MANIQA$\uparrow$ \\
        \midrule
        w/o skip connection & 30.52 & 0.952 & \textbf{53.98} & \textbf{0.6029} \\
        w/ skip connection (ours) & \textbf{33.05} & \textbf{0.963} & 50.81 & 0.5668 \\
        \bottomrule
    \end{tabular}
    }
    \label{tab:skip_connection_ablation}
    \vspace{-0.2cm}
\end{table}

\clearpage
\section{More Qualitative Results}\label{sec:app_visual_results}
\vspace{-0.3cm}
\begin{figure}[!ht]
  \centering
\begin{minipage}[!ht]{.48\linewidth}
  \centering
  \centerline{\includegraphics[width=6.7cm]{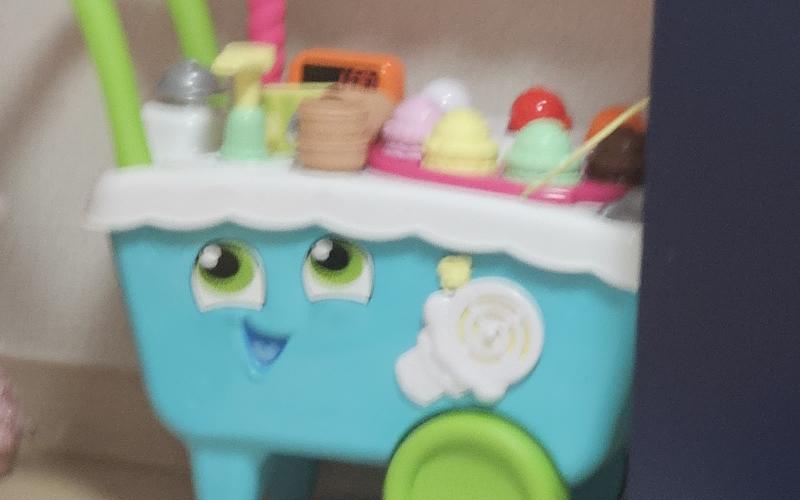}}
  \centerline{(a) Blur Input}\medskip
  \centerline{\includegraphics[width=6.7cm]{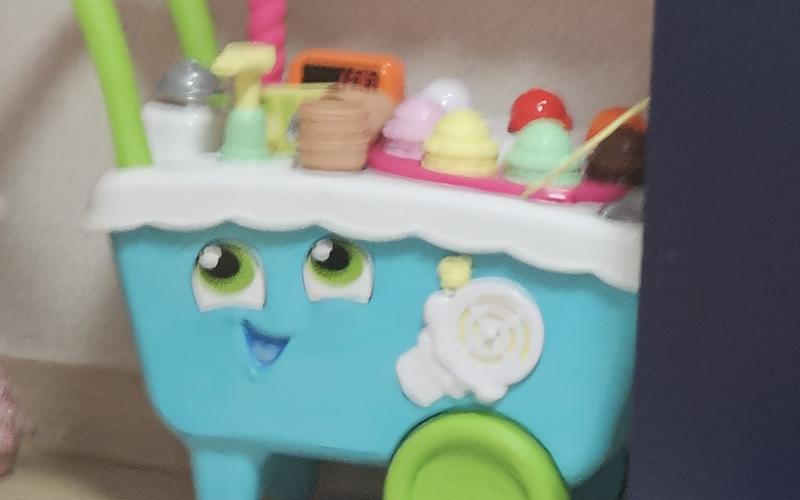}}
  \centerline{(c) DiffIR~\cite{diffir}}\medskip
\end{minipage}
\begin{minipage}[!ht]{.48\linewidth}
  \centering
  \centerline{\includegraphics[width=6.7cm]{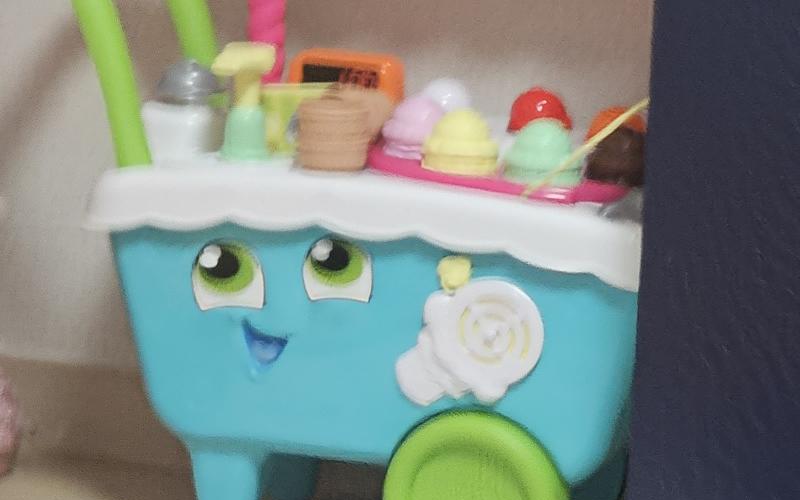}}
  \centerline{(b) NAFNet~\cite{nafnet}}\medskip
  \centerline{\includegraphics[width=6.7cm]{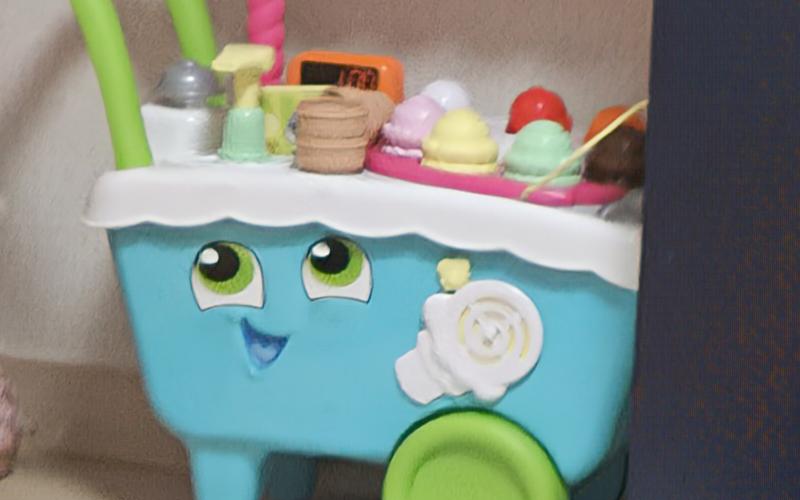}}
  \centerline{(d) DeblurFlow (ours)}\medskip
\end{minipage}
\vspace{-0.3cm}
\end{figure}

\begin{figure}[!ht]
  \centering
\begin{minipage}[!ht]{.48\linewidth}
  \centering
  \centerline{\includegraphics[width=6.7cm]{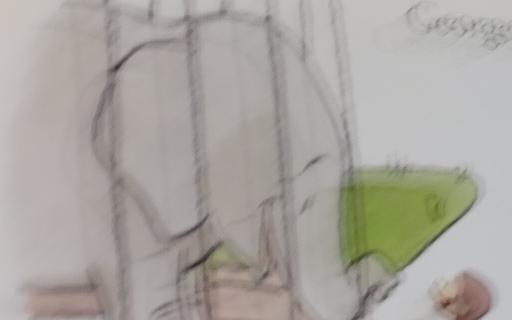}}
  \centerline{(a) Blur Input}\medskip
  \centerline{\includegraphics[width=6.7cm]{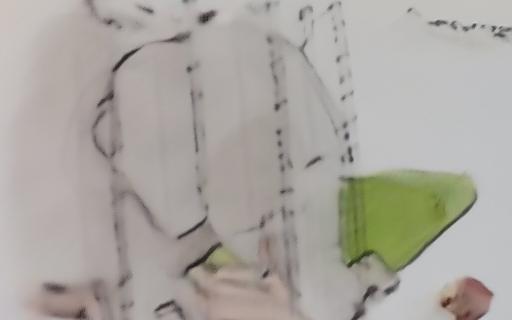}}
  \centerline{(c) DiffIR~\cite{diffir}}\medskip
\end{minipage}
\begin{minipage}[!ht]{.48\linewidth}
  \centering
  \centerline{\includegraphics[width=6.7cm]{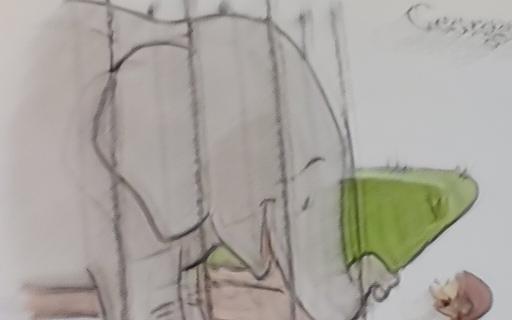}}
  \centerline{(b) NAFNet~\cite{nafnet}}\medskip
  \centerline{\includegraphics[width=6.7cm]{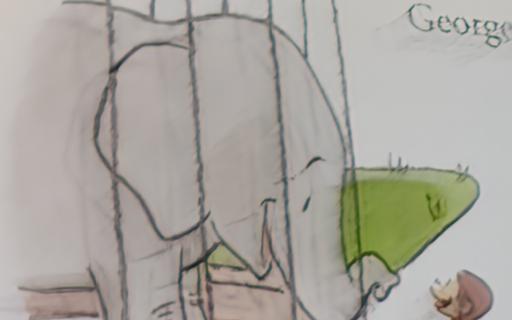}}
  \centerline{(d) DeblurFlow (ours)}\medskip
\end{minipage}
\end{figure}

\clearpage
\begin{figure}[!ht]
  \centering
\begin{minipage}[!ht]{.48\linewidth}
  \centering
  \centerline{\includegraphics[width=6.7cm]{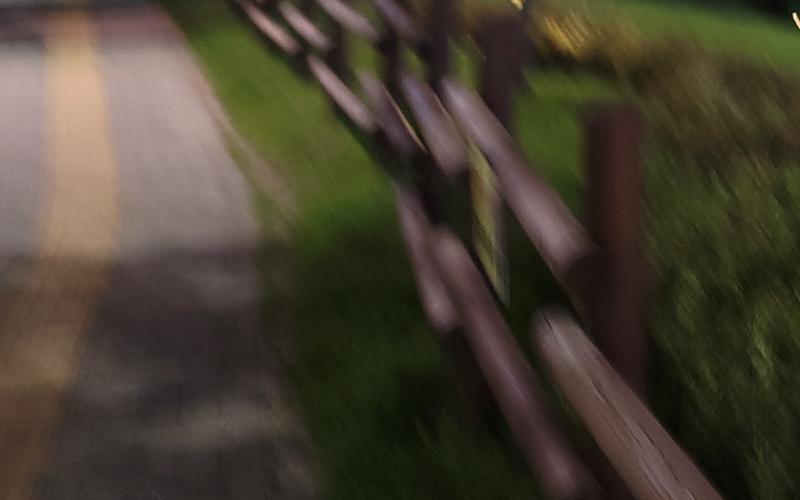}}
  \centerline{(a) Blur Input}\medskip
  \centerline{\includegraphics[width=6.7cm]{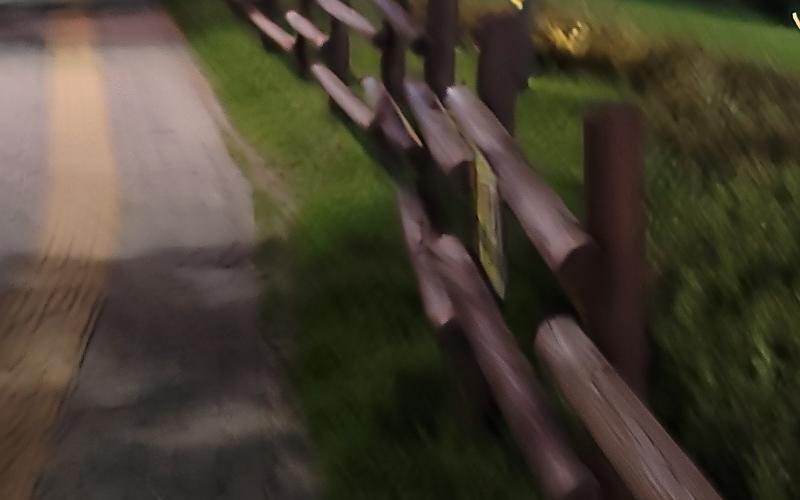}}
  \centerline{(c) DiffIR~\cite{diffir}}\medskip
\end{minipage}
\begin{minipage}[!ht]{.48\linewidth}
  \centering
  \centerline{\includegraphics[width=6.7cm]{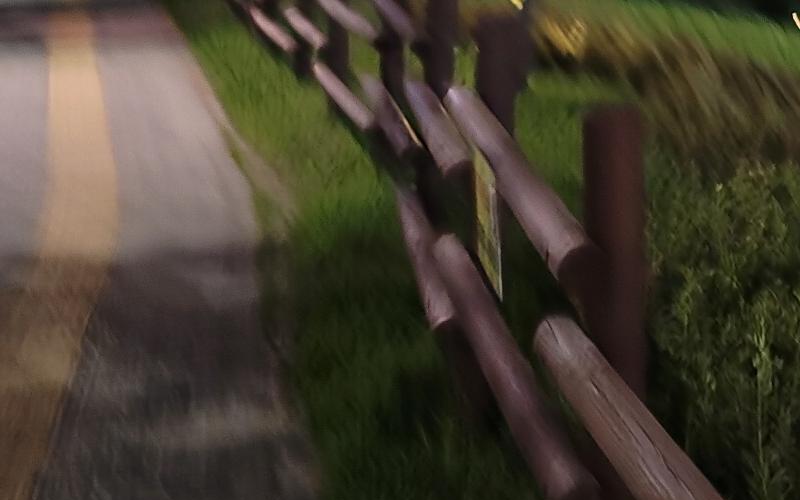}}
  \centerline{(b) NAFNet~\cite{nafnet}}\medskip
  \centerline{\includegraphics[width=6.7cm]{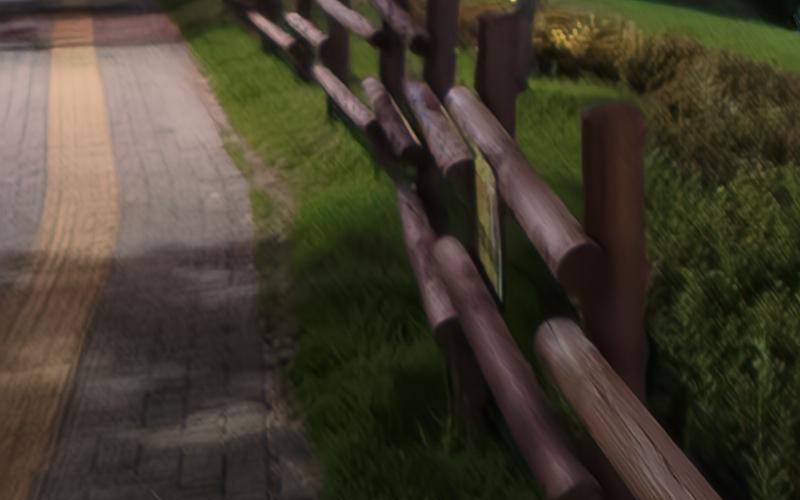}}
  \centerline{(d) DeblurFlow (ours)}\medskip
\end{minipage}
\vspace{-0.3cm}
\end{figure}

\begin{figure}[!ht]
  \centering
\begin{minipage}[!ht]{.48\linewidth}
  \centering
  \centerline{\includegraphics[width=6.7cm]{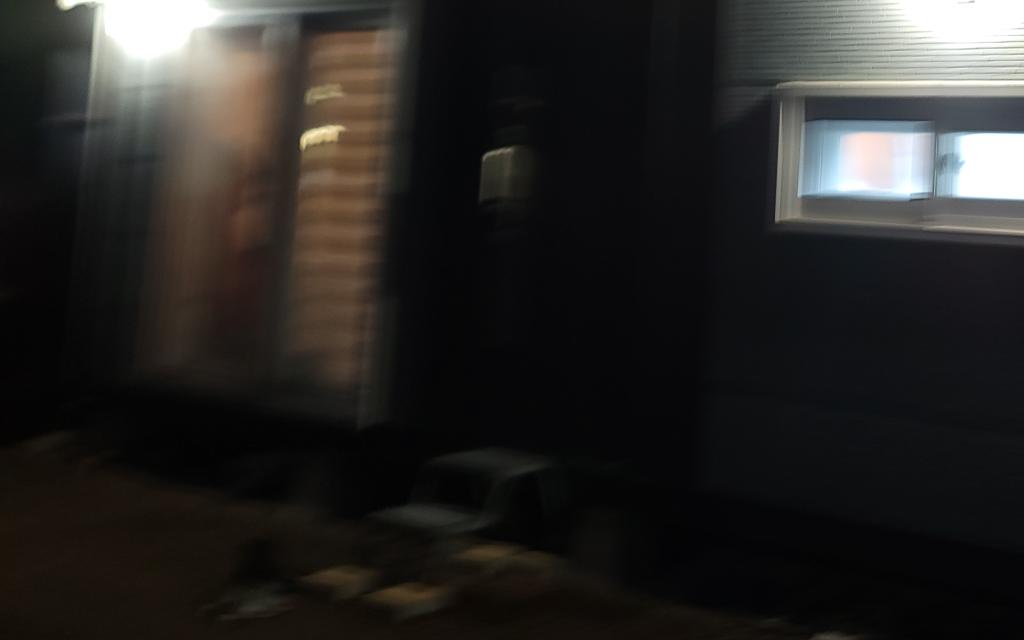}}
  \centerline{(a) Blur Input}\medskip
  \centerline{\includegraphics[width=6.7cm]{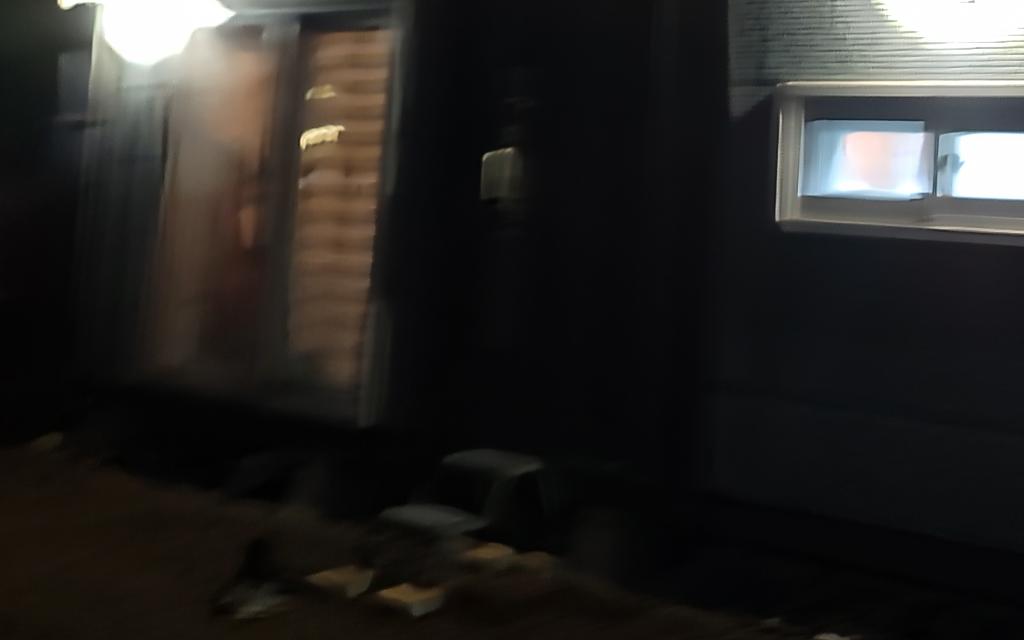}}
  \centerline{(c) DiffIR~\cite{diffir}}\medskip
\end{minipage}
\begin{minipage}[!ht]{.48\linewidth}
  \centering
  \centerline{\includegraphics[width=6.7cm]{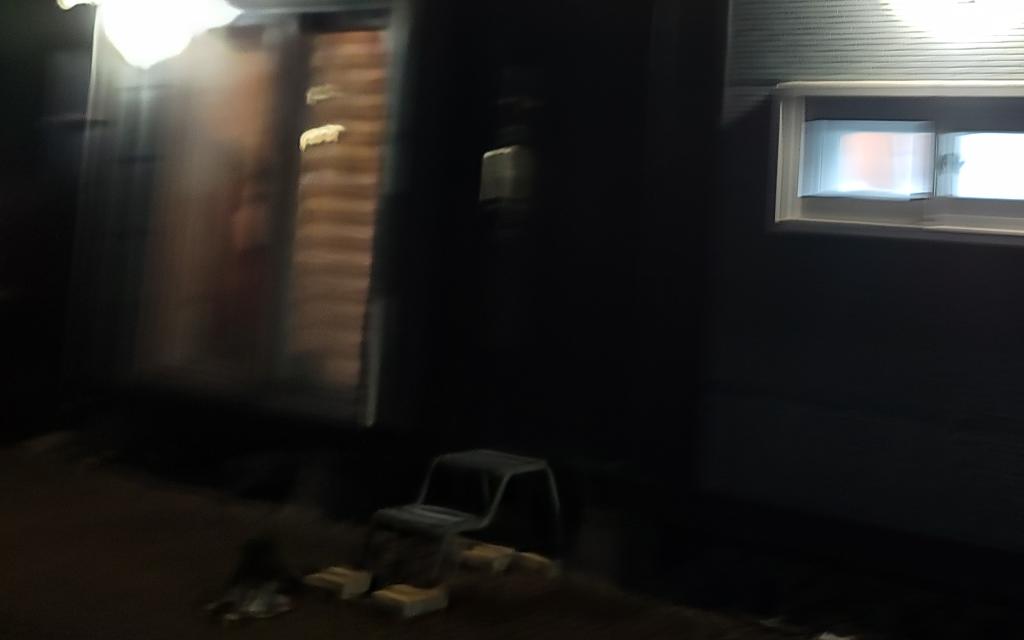}}
  \centerline{(b) NAFNet~\cite{nafnet}}\medskip
  \centerline{\includegraphics[width=6.7cm]{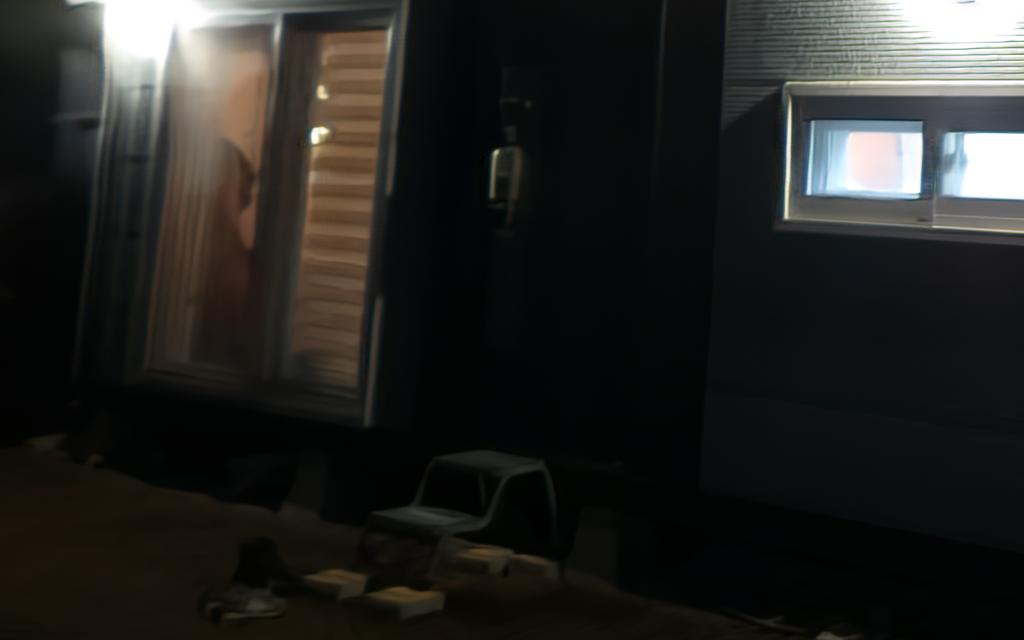}}
  \centerline{(d) DeblurFlow (ours)}\medskip
\end{minipage}
\vspace{-0.3cm}
\caption{Visual comparison results on real-world blur images.}
\vspace{-0.3cm}
\label{fig:app_realworld}
\end{figure}

\begin{figure}[!ht]
\begin{minipage}[!t]{.32\linewidth}
  \centering
  \centerline{\includegraphics[width=4.5cm]{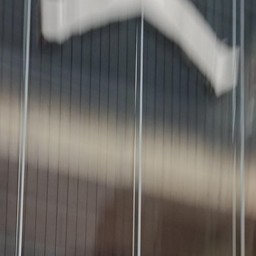}}
  \centerline{\small{(a) Blur}}\medskip
\end{minipage}
\begin{minipage}[!ht]{.32\linewidth}
  \centering
  \centerline{\includegraphics[width=4.5cm]{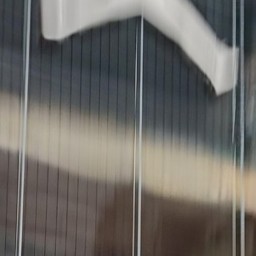}}
  \centerline{\small{(b) NAFNet~\cite{nafnet}}}\medskip
\end{minipage}
\begin{minipage}[!ht]{.32\linewidth}
  \centering
  \centerline{\includegraphics[width=4.5cm]{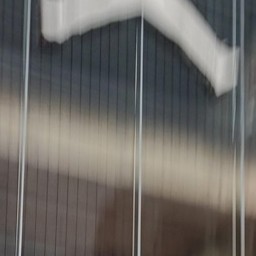}}
  \centerline{\small{(c) FFTFormer~\cite{fftformer}}}\medskip
\end{minipage}
\end{figure}
\begin{figure}[!ht]
\begin{minipage}[!ht]{.32\linewidth}
  \centering
  \centerline{\includegraphics[width=4.5cm]{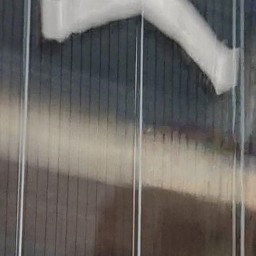}}
  \centerline{\small{(d) IR-SDE~\cite{irsde}}}\medskip
\end{minipage}
\begin{minipage}[!ht]{.32\linewidth}
  \centering
  \centerline{\includegraphics[width=4.5cm]{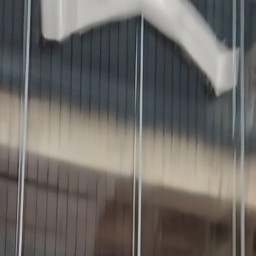}}
  \centerline{\small{(e) DiffIR~\cite{diffir}}}\medskip
\end{minipage}
\begin{minipage}[!ht]{.32\linewidth}
  \centering
  \centerline{\includegraphics[width=4.5cm]{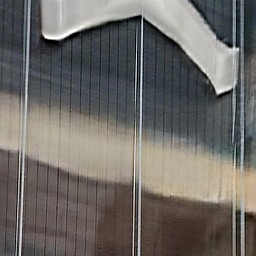}}
  \centerline{\small{(f) DeblurFlow (ours)}}\medskip
\end{minipage}
\end{figure}
\begin{figure}[!ht]
\begin{minipage}[!ht]{.32\linewidth}
  \centering
  \centerline{\includegraphics[width=4.5cm]{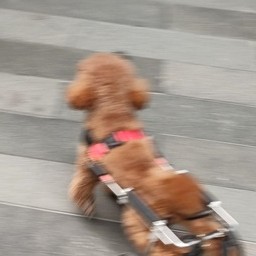}}
  \centerline{\small{(a) Blur}}\medskip
\end{minipage}
\begin{minipage}[!ht]{.32\linewidth}
  \centering
  \centerline{\includegraphics[width=4.5cm]{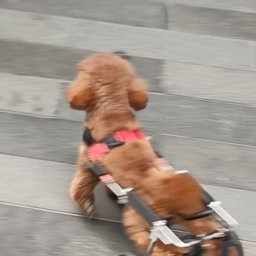}}
  \centerline{\small{(b) NAFNet~\cite{nafnet}}}\medskip
\end{minipage}
\begin{minipage}[!ht]{.32\linewidth}
  \centering
  \centerline{\includegraphics[width=4.5cm]{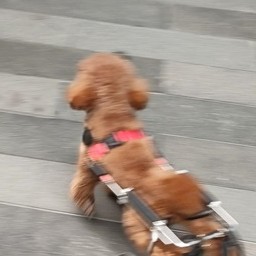}}
  \centerline{\small{(c) FFTFormer~\cite{fftformer}}}\medskip
\end{minipage}
\end{figure}
\begin{figure}[!ht]
\begin{minipage}[!ht]{.32\linewidth}
  \centering
  \centerline{\includegraphics[width=4.5cm]{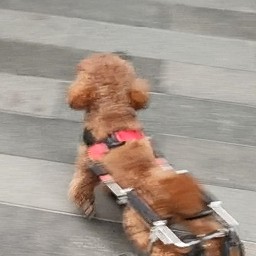}}
  \centerline{\small{(d) IR-SDE~\cite{irsde}}}\medskip
\end{minipage}
\begin{minipage}[!ht]{.32\linewidth}
  \centering
  \centerline{\includegraphics[width=4.5cm]{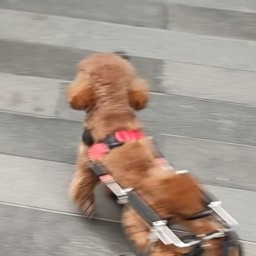}}
  \centerline{\small{(e) DiffIR~\cite{diffir}}}\medskip
\end{minipage}
\begin{minipage}[!ht]{.32\linewidth}
  \centering
  \centerline{\includegraphics[width=4.5cm]{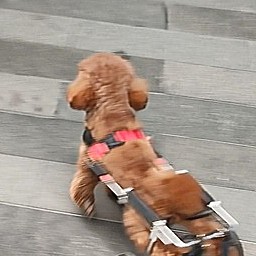}}
  \centerline{\small{(f) DeblurFlow (ours)}}\medskip
\end{minipage}
\vspace{-0.3cm}
\end{figure}

\begin{figure}[!ht]
\begin{minipage}[!ht]{.32\linewidth}
  \centering
  \centerline{\includegraphics[width=4.5cm]{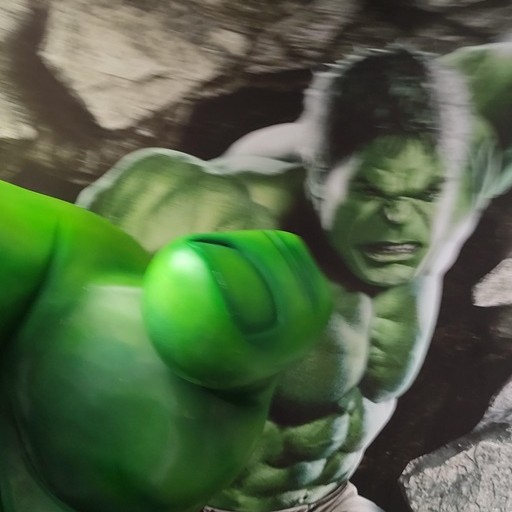}}
  \centerline{\small{(a) Blur}}\medskip
\end{minipage}
\begin{minipage}[!ht]{.32\linewidth}
  \centering
  \centerline{\includegraphics[width=4.5cm]{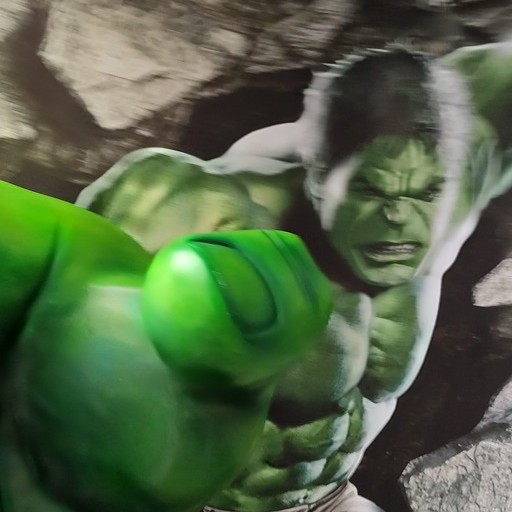}}
  \centerline{\small{(b) NAFNet~\cite{nafnet}}}\medskip
\end{minipage}
\begin{minipage}[!ht]{.32\linewidth}
  \centering
  \centerline{\includegraphics[width=4.5cm]{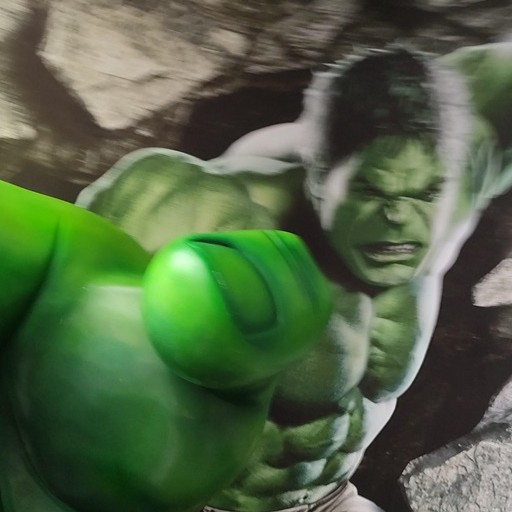}}
  \centerline{\small{(c) FFTFormer~\cite{fftformer}}}\medskip
\end{minipage}
\end{figure}
\begin{figure}[!ht]
\begin{minipage}[!ht]{.32\linewidth}
  \centering
  \centerline{\includegraphics[width=4.5cm]{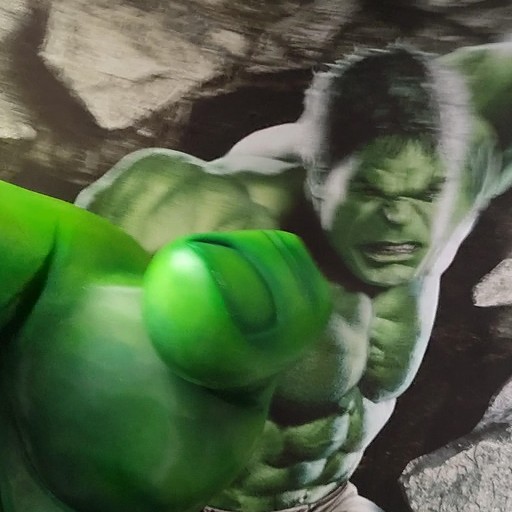}}
  \centerline{\small{(d) IR-SDE~\cite{irsde}}}\medskip
\end{minipage}
\begin{minipage}[!ht]{.32\linewidth}
  \centering
  \centerline{\includegraphics[width=4.5cm]{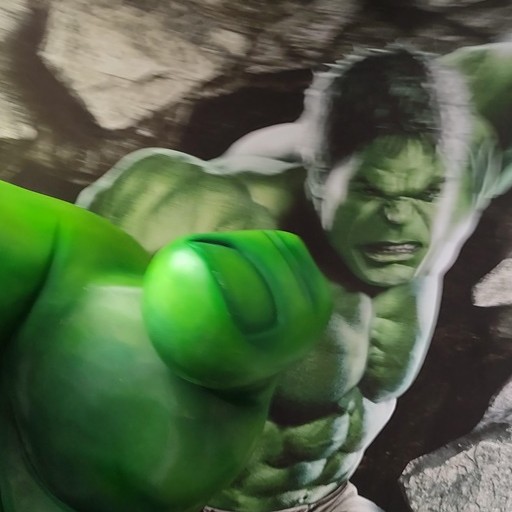}}
  \centerline{\small{(e) DiffIR~\cite{diffir}}}\medskip
\end{minipage}
\begin{minipage}[!ht]{.32\linewidth}
  \centering
  \centerline{\includegraphics[width=4.5cm]{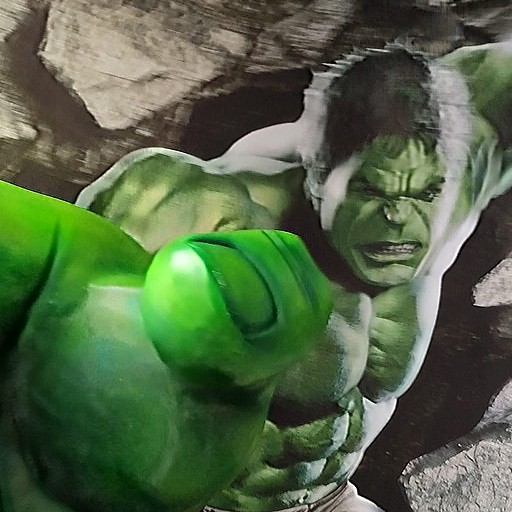}}
  \centerline{\small{(f) DeblurFlow (ours)}}\medskip
\end{minipage}
\end{figure}
\begin{figure}[!ht]
\begin{minipage}[!ht]{.32\linewidth}
  \centering
  \centerline{\includegraphics[width=4.5cm]{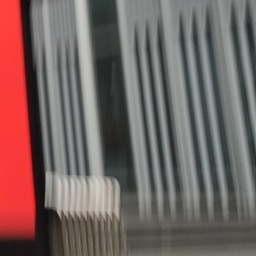}}
  \centerline{\small{(a) Blur}}\medskip
\end{minipage}
\begin{minipage}[!ht]{.32\linewidth}
  \centering
  \centerline{\includegraphics[width=4.5cm]{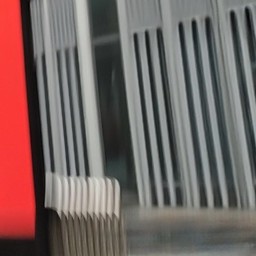}}
  \centerline{\small{(b) NAFNet~\cite{nafnet}}}\medskip
\end{minipage}
\begin{minipage}[!ht]{.32\linewidth}
  \centering
  \centerline{\includegraphics[width=4.5cm]{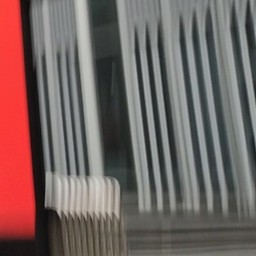}}
  \centerline{\small{(c) FFTFormer~\cite{fftformer}}}\medskip
\end{minipage}
\end{figure}

\begin{figure}[!ht]
\begin{minipage}[!ht]{.32\linewidth}
  \centering
  \centerline{\includegraphics[width=4.5cm]{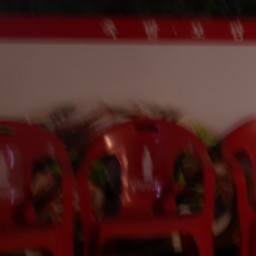}}
  \centerline{\small{(a) Blur}}\medskip
\end{minipage}
\begin{minipage}[!ht]{.32\linewidth}
  \centering
  \centerline{\includegraphics[width=4.5cm]{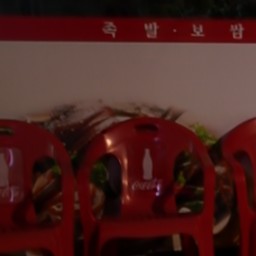}}
  \centerline{\small{(b) NAFNet~\cite{nafnet}}}\medskip
\end{minipage}
\begin{minipage}[!ht]{.32\linewidth}
  \centering
  \centerline{\includegraphics[width=4.5cm]{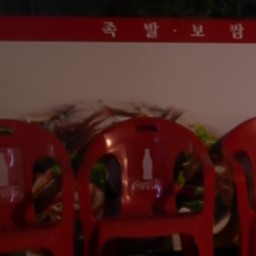}}
  \centerline{\small{(c) Hi-Diff~\cite{hidiff}}}\medskip
\end{minipage}
\end{figure}
\begin{figure}[!ht]
\begin{minipage}[!ht]{.32\linewidth}
  \centering
  \centerline{\includegraphics[width=4.5cm]{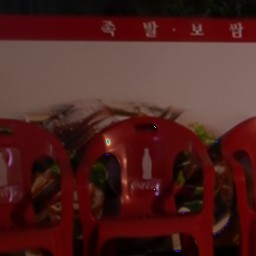}}
  \centerline{\small{(d) DiffIR~\cite{diffir}}}\medskip
\end{minipage}
\begin{minipage}[!ht]{.32\linewidth}
  \centering
  \centerline{\includegraphics[width=4.5cm]{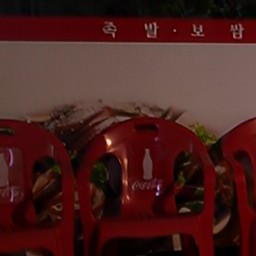}}
  \centerline{\small{(e) DeblurFlow (ours)}}\medskip
\end{minipage}
\begin{minipage}[!ht]{.32\linewidth}
  \centering
  \centerline{\includegraphics[width=4.5cm]{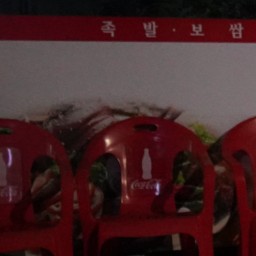}}
  \centerline{\small{(f) Ground-Truth}}\medskip
\end{minipage}
\end{figure}
\begin{figure}[!ht]
\begin{minipage}[!ht]{.32\linewidth}
  \centering
  \centerline{\includegraphics[width=4.5cm]{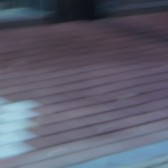}}
  \centerline{\small{(a) Blur}}\medskip
\end{minipage}
\begin{minipage}[!ht]{.32\linewidth}
  \centering
  \centerline{\includegraphics[width=4.5cm]{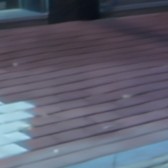}}
  \centerline{\small{(b) NAFNet~\cite{nafnet}}}\medskip
\end{minipage}
\begin{minipage}[!ht]{.32\linewidth}
  \centering
  \centerline{\includegraphics[width=4.5cm]{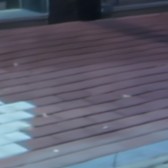}}
  \centerline{\small{(c) Hi-Diff~\cite{hidiff}}}\medskip
\end{minipage}
\end{figure}
\begin{figure}[!ht]
\begin{minipage}[!ht]{.32\linewidth}
  \centering
  \centerline{\includegraphics[width=4.5cm]{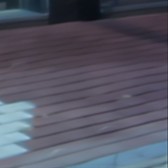}}
  \centerline{\small{(d) DiffIR~\cite{diffir}}}\medskip
\end{minipage}
\begin{minipage}[!ht]{.32\linewidth}
  \centering
  \centerline{\includegraphics[width=4.5cm]{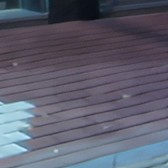}}
  \centerline{\small{(e) DeblurFlow (ours)}}\medskip
\end{minipage}
\begin{minipage}[!ht]{.32\linewidth}
  \centering
  \centerline{\includegraphics[width=4.5cm]{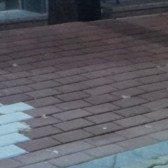}}
  \centerline{\small{(f) Ground-Truth}}\medskip
\end{minipage}
\vspace{-0.3cm}
\end{figure}

\begin{figure}[!ht]
\begin{minipage}[!ht]{.32\linewidth}
  \centering
  \centerline{\includegraphics[width=4.5cm]{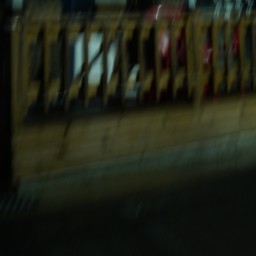}}
  \centerline{\small{(a) Blur}}\medskip
\end{minipage}
\begin{minipage}[!ht]{.32\linewidth}
  \centering
  \centerline{\includegraphics[width=4.5cm]{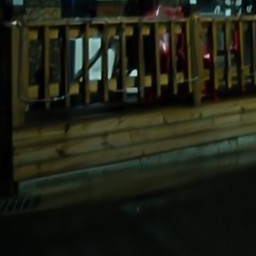}}
  \centerline{\small{(b) NAFNet~\cite{nafnet}}}\medskip
\end{minipage}
\begin{minipage}[!ht]{.32\linewidth}
  \centering
  \centerline{\includegraphics[width=4.5cm]{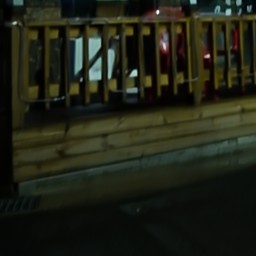}}
  \centerline{\small{(c) Hi-Diff~\cite{hidiff}}}\medskip
\end{minipage}
\end{figure}
\begin{figure}[!ht]
\begin{minipage}[!ht]{.32\linewidth}
  \centering
  \centerline{\includegraphics[width=4.5cm]{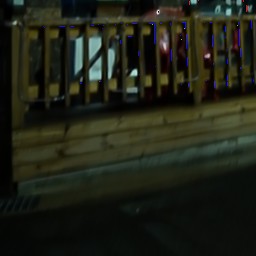}}
  \centerline{\small{(d) DiffIR~\cite{diffir}}}\medskip
\end{minipage}
\begin{minipage}[!ht]{.32\linewidth}
  \centering
  \centerline{\includegraphics[width=4.5cm]{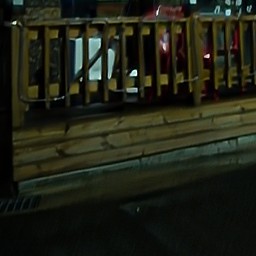}}
  \centerline{\small{(e) DeblurFlow (ours)}}\medskip
\end{minipage}
\begin{minipage}[!ht]{.32\linewidth}
  \centering
  \centerline{\includegraphics[width=4.5cm]{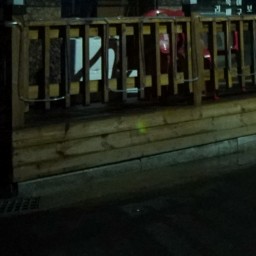}}
  \centerline{\small{(f) Ground-Truth}}\medskip
\end{minipage}
\end{figure}
\begin{figure}[!ht]
\begin{minipage}[!ht]{.32\linewidth}
  \centering
  \centerline{\includegraphics[width=4.5cm]{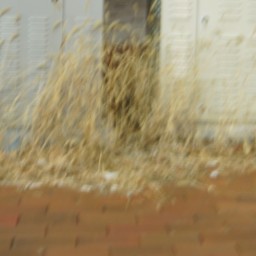}}
  \centerline{\small{(a) Blur}}\medskip
\end{minipage}
\begin{minipage}[!ht]{.32\linewidth}
  \centering
  \centerline{\includegraphics[width=4.5cm]{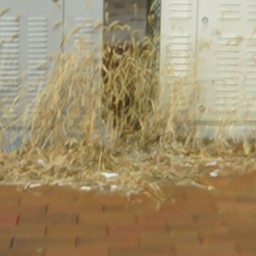}}
  \centerline{\small{(b) NAFNet~\cite{nafnet}}}\medskip
\end{minipage}
\begin{minipage}[!ht]{.32\linewidth}
  \centering
  \centerline{\includegraphics[width=4.5cm]{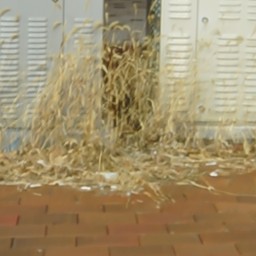}}
  \centerline{\small{(c) Hi-Diff~\cite{hidiff}}}\medskip
\end{minipage}
\end{figure}
\begin{figure}[!ht]
\begin{minipage}[!ht]{.32\linewidth}
  \centering
  \centerline{\includegraphics[width=4.5cm]{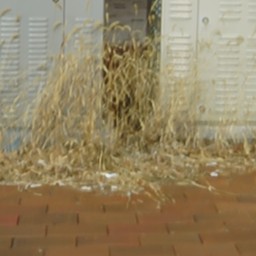}}
  \centerline{\small{(d) DiffIR~\cite{diffir}}}\medskip
\end{minipage}
\begin{minipage}[!ht]{.32\linewidth}
  \centering
  \centerline{\includegraphics[width=4.5cm]{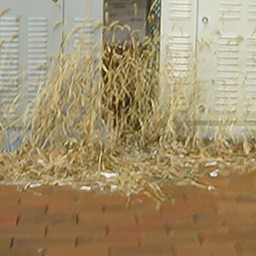}}
  \centerline{\small{(e) DeblurFlow (ours)}}\medskip
\end{minipage}
\begin{minipage}[!ht]{.32\linewidth}
  \centering
  \centerline{\includegraphics[width=4.5cm]{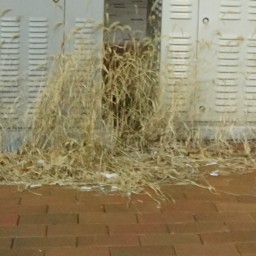}}
  \centerline{\small{(f) Ground-Truth}}\medskip
\end{minipage}
\vspace{-0.3cm}
\caption{Visual comparison results on RealBlur-J~\cite{realblur}.}
\vspace{-0.3cm}
\label{fig:app_result_realblur}
\end{figure}


\end{document}